\documentclass[preprint,12pt,authoryear]{elsarticle}

\usepackage{amssymb}
\usepackage[utf8]{inputenc}
\usepackage{amsfonts,amsmath,amssymb}
\usepackage{booktabs}
\usepackage[ruled,vlined]{algorithm2e}
\usepackage{amsmath}
\usepackage{hyperref}
\usepackage{capt-of}
\usepackage{subcaption}
\usepackage{comment}
\usepackage{booktabs}
\usepackage{verbatim}
\usepackage{makecell}
\usepackage{graphicx}
\usepackage{subcaption}
\usepackage{multirow}
\usepackage{algpseudocode}
\usepackage{geometry}
\usepackage{adjustbox}
\usepackage{changepage} 
\usepackage[hypcap=false]{caption}
\usepackage{rotating}      
\usepackage{xcolor}

\journal{Pattern Recognition}

\begin{document}

\begin{frontmatter}



\title{TriSig: Assessing the statistical significance of triclusters}


\author[inst1,inst2,inst3]{Leonardo Alexandre\corref{cor1}}
\ead{leonardoalexandre@tecnico.ulisboa.pt}
\cortext[cor1]{Leonardo Alexandre}
\author[inst3]{Rafael S. Costa}
\author[inst1,inst2]{Rui Henriques\corref{cor2}}
\ead{rmch@tecnico.ulisboa.pt}
\cortext[cor2]{Rui Henriques}

\affiliation[inst1]{organization={INESC-ID, Lisboa, Portugal},
            addressline={R. Alves Redol 9, 1000-029, Lisboa}, 
            country={Portugal}}
\affiliation[inst2]{organization={Instituto Superior Técnico, University of Lisbon, Lisbon},
    country={Portugal}}
\affiliation[inst3]{organization={LAQV-REQUIMTE, Department of Chemistry, NOVA School of Science and Technology, Universidade NOVA de Lisboa},
            addressline={Campus Caparica, 2829-516, Caparica}, 
            country={Portugal}}

\begin{abstract}
Tensor data analysis allows researchers to uncover novel patterns and relationships that cannot be obtained from matrix data alone. The information inferred from the patterns provides valuable insights into disease progression, bioproduction processes, weather fluctuations, and group dynamics. However, spurious and redundant patterns hamper this process. This work aims at proposing a statistical frame to assess the probability of patterns in tensor data to deviate from null expectations, extending well-established principles for assessing the statistical significance of patterns in matrix data. A comprehensive discussion on binomial testing for false positive discoveries is entailed at the light of: variable dependencies, temporal dependencies and misalignments, and \textit{p}-value corrections under the Benjamini-Hochberg procedure. Results gathered from the application of state-of-the-art triclustering algorithms over distinct real-world case studies in biochemical and biotechnological domains confer validity to the proposed statistical frame while revealing vulnerabilities of some triclustering searches. The proposed assessment can be incorporated into existing triclustering algorithms to mitigate false positive/spurious discoveries and further prune the search space, reducing their computational complexity. 

 
\vskip 0.2cm 

\noindent\textbf{Availability:} The code is freely available at \href{https://github.com/JupitersMight/TriSig}{https://github.com/JupitersMight/ TriSig} under the MIT license.

\end{abstract}

\begin{keyword}
Triclustering \sep Tensor data \sep Pattern discovery \sep Statistical significance \sep Temporal pattern mining \sep Multivariate time series data 



\end{keyword}

\end{frontmatter}


\section{Introduction}

In recent years, the increased capacity for comprehensively monitoring systems' behavior led to the emergence of tensor data structures such as three-dimensional data, also known as cubic or three-way data \citep{henriques2018triclustering}. The nature of tensor data allows for a more comprehensive understanding of the underlying processes and relationships, proving to be a valuable source for making informed decisions and predictions. Some real-world examples in the biological domain include omic data to study how treatments affect a given tissue \citep{kanno2006per, cardoso2019gene, strober2019dynamic, white2020reference}, improve the effectiveness of plantations against drought \citep{groen2020strength}, understand development patterns \citep{liu2020inter}, amongst other ends \citep{yalccin2020analysis, kim2019long, mandal2022poptric, gan2018tri}. In social studies, user behavior has been monitored to better understand group dynamics such as user preferences \citep{gnatyshak2012gaining, gnatyshak2014greedy, ignatov2015triadic} and interactions \citep{joo2019towards, song2019triadic, ahn2022bifold, alessandroni2020musical}. Geophysical data has been considered to better understand weather patterns \citep{wu2020overview, kazemi2021generalized}, seismic activity \citep{amaro2021generating}, or the fluctuation of the population's needs \citep{melgar2022new, guigoures2018discovering, kremser2021southern}. Finally, in the biomedical field, pattern discovery from health records has been considered for personalized medicine \citep{alexandre2021mining}, understanding diseases \citep{soares2020towards}, and neuroscience advances \citep{rahaman2022tri}.

Although many pattern discovery approaches are available for actionable knowledge acquisition \citep{oberski2016mixture, nagin2010group, wu2013longitudinal, jung2008introduction}, we focus on triclustering \citep{henriques2018triclustering}, a well-established unsupervised task for extracting patterns within tensor data. Triclustering belongs to the same branch of unsupervised tasks as clustering and biclustering. But contrary to clustering, where clusters represent groups of observations meaningfully correlated along the overall feature space, and biclustering, only prepared to handle matricial data, triclustering can make full use of the tensor data properties. Patterns extracted by triclustering algorithms, referred to as triclusters, are defined by a set of observations meaningfully correlated on a subset of variables along a subset of contexts. 
Triclusters can represent a coherent temporal progression of specific features for a group of observations, meaning that features can vary throughout time as long as the desirable form of correlation is preserved. Triclusters should also meet rigorous dissimilarity and statistical significance criteria \citep{henriques2018triclustering}. Dissimilarity criteria prevent redundant triclusters from forming, while statistical significance ensures that the probability of the tricluster occurring against null expectations is unexpectedly low. It should be noted that pursuing a stronger correlation between elements of a tricluster is not sufficient to guarantee its statistical significance. Triclusters with a small number of elements can yield high correlation levels that still occur by chance. Similarly, domain relevance (e.g., functional enrichment against knowledge bases \citep{ashburner2000gene}) is, alone, insufficient to ensure statistical significance. 

\cite{henriques2018triclustering} provide an entry point to the state-of-the-art research of tensor data using triclustering, covering former studies considering statistical significance assessments \citep{moise2008finding, sim2010discovering}. \cite{moise2008finding} placed a uniform null assumption to assess whether the volume of a given tricluster deviates from expectations under a Binomial test. Applicability is constrained to identically distributed variables. \cite{sim2010discovering} proposed a correlation metric based on mutual information, termed CI, to identify statistically significant triclusters. A statistically significant tricluster yields an unexpectedly high CI (rarely co-occurring values), with the CI distribution being modeled under a gamma or Weibull distribution. Whilst not many, some works have proposed and discussed the statistical significance of triclusters after the survey was published. \cite{uvzupyte2020test} test the statistical significance by using Hotelling's T-squared test, a multivariate generalization of the $t$-student test where they test if the means for a group equals a hypothetical vector of means, under an exponential assumption. \cite{biswal2020trirnsc} proposed a novel triclustering algorithm, where extracted triclusters are assessed by three distinct metrics: a triclustering quality index based square residue (MSR) \citep{bhar2013coexpression}; a statistical difference from the background (SDB) that ensures that the triclusters yields differential values against background regularities;  
and a functional enrichment $p$-value (probability of intersection) returned by the GO Term \textsf{findertool} \citep{ashburner2000gene}. Please note that domain relevance (under, for instance, a functional enrichment analysis) and statistical significance are distinct stances on a tricluster's interestingness that may be contradictory \citep{henriques2018bsig}. \cite{williams2022modelling} use Structural equation modeling (SEM), a statistical technique to model complex relationships between variables, 
to assess how well the observed data fit a learnt SEM-based model at two levels: 
the statistical significance of latent factors (described by a set of variables) under a t-statistic, and the statistical significance of the model under a $\chi^2$ test.

Despite their relevance, to the best of our knowledge, current state-of-the-art triclustering approaches fail to provide a generalized assessment of a pattern's statistical significance (i.e., whether a pattern occurs by chance or not against a null data model) that takes into consideration the unique regularities within tensor data (e.g., variable domains and dependencies). In light of this, we propose an extended frame, expanding well-established principles for assessing the statistical significance of biclustering solutions. Our solution is able to accommodate different homogeneity criteria, data types, variable dependencies, temporal properties (e.g., contiguity and misalignments), and, consistently with \cite{sim2010discovering}, \textit{p}-value corrections. 

In this context, this study presents the first comprehensive statistical frame to mitigate false positive discoveries from tensor data with varying properties. The proposed methodology is tested using four case studies and synthetic tensor data. The application of state-of-the-art triclustering algorithms unravel behavioral vulnerabilities and further show how statistical significance varies depending on the pattern type and size, background data, dependency assumptions, and applicable corrections. The proposed methodology is able to approximate multivariate boundaries that can be used to separate spurious discoveries from statistically significant ones. 



The rest of this paper is organized as follows: the remainder of this section provides the essential background (for more detailed formulations, the reader can check \cite{henriques2018triclustering}). Section 2 introduces the proposed methodology. Section 3 discusses the gathered results from real-world and synthetic data. Concluding remarks and implications are finally drawn.

\subsection{Background}

Let a two-dimensional dataset (matrix data), \textbf{A}, be defined by $n$ observations ($X = \{x_1, ..., x_n\}$) and $m$ variables ($Y = \{y_1, ..., y_m\}$), with $n \times m$ elements $a_{ij}$. \textit{Clustering} and \textit{biclustering} tasks aim to extract \textit{clusters} 
$I_i  \subseteq X$ or \textit{biclusters} $B_k = (I_k, J_k)$, each given by a subset of observations, $I_k  \subseteq X$, correlated on a subset of variables, $J_k \subseteq Y$. \textbf{Homogeneity} criteria, commonly guaranteed through the use of a merit function, guides the formation of (bi)clusters 
\citep{madeira2004biclustering}. Given a bicluster, its values $a_{ij}=c_j+\gamma_i+\eta_{ij}$ 
can be described by value expectations $c_j$, adjustments $\gamma_i$, and noise $\eta_{ij}$, then the \textbf{bicluster pattern} $\varphi_{B}$ is the ordered set of values in the absence of adjustments and noise, $\varphi_{B} =\{ c_{j} | y_j \in J\}$. 
In addition to homogeneity criteria, \textbf{statistical significance} criteria \citep{henriques2018bsig} guarantee that the probability of a bicluster's occurrence (against a null model) deviates from expectations. Furthermore, \textbf{dissimilarity} criteria \citep{henriques2017bicpams} 
can be placed to further guarantee the absence of redundant biclusters.



A three-dimensional dataset (three-way tensor data), \textbf{A}, is defined by $n$ observations $X = \{x_1, ..., x_n\}$, $m$ variables $Y = \{y_1, ..., y_m\}$, and $p$ contexts $Z = \{z_1, ..., z_p\}$. Elements $a_{ijk}$ relate observation $x_i$, attribute $y_j$, and context $z_k$. Identical to two-dimensional data, three-dimensional data can be \textit{real-valued} $(a_{ijk} \in \mathbb{R})$, symbolic ($a_{ijk} \in \Sigma$, where $\Sigma$ is a set of nominal or ordinal symbols), integer ($a_{ijk} \in \mathbb{Z}$), or \textit{non-identically distributed} ($a_{ijk} \in \mathcal{A}_j$, where $\mathcal{A}_j$ is the domain of $y_j$'s variable), and contain missing elements, $a_{ijk} \in \mathcal{A}_j \cup \emptyset$. When the context dimension corresponds to a set of time points ($Z = \{t_1, t_2, ..., t_p\}$), we are in the presence of a temporal three-dimensional dataset, also referred as $m$-order multivariate time series data. 

Given three-way tensor data \textbf{A} with $n$ observations, $m$ variables, and $p$ contexts/time points, a \textbf{tricluster} $T = (I, J, K)$ is a subspace of the original space, where $I \subseteq X$, $J \subseteq Y$, and $K \subseteq Z$ are subsets of observations, variables, and contexts/time points, respectively \citep{henriques2018triclustering}. 
The \textbf{triclustering task} task aims to find a set of triclusters $\mathcal{T} = \{T_1, ..., T_l\}$ such that each tricluster $T_i$ satisfies specific criteria of homogeneity, dissimilarity, and statistical significance. 
Let each element of a tricluster, $a_{ijk}$, be described by a base value $c_{jk}$, observation adjustment $\gamma_i$ 
and noise $\eta_{ijk}$. The tricluster \textbf{pattern}, $\varphi_{T}$, is an ordered set of value expectations along the subset of variables and context dimensions in the absence of adjustments and noise: $\varphi_{T} = \{c_{jk} | y_j \in J, z_k \in K, c_{jk} \in Y_k\}$.  

\textbf{Homogeneity} criteria determine the structure, coherence, and quality of a triclustering solution \citep{henriques2018triclustering}, where: 1) the \textit{structure} is described by the number, size, shape, and position of triclusters, 2) the \textit{coherence} of a tricluster is defined by the observed correlation of values (coherence assumption) and the allowed deviation from expectations (coherence strength), and 3) the \textit{quality} of a tricluster is defined by the type and amount of tolerated noise. A tricluster has \textbf{constant coherence} when the subspace exhibits constant (for symbolic data) or approximately constant (real-valued data) values. \textbf{Dissimilarity} criteria guarantee that any tricluster similar to another tricluster with higher priority is removed from $\mathcal{T}$ and (possibly) used to refine similar triclusters in $\mathcal{T}$. A tricluster $T = (I, J, K)$ is \textit{maximal} if and only if there is no other tricluster $(\textbf{I',J',K'})$ such that $I \subseteq I' \wedge J \subseteq J' \wedge K \subseteq K'$ satisfying the given criteria. 
Finally, and foremost, \textbf{Statistical significance criteria} guarantee that the probability of each retrieved tricluster occurring against a null data model is unexpectedly low.

\section{Methodology}

The proposed methodology offers a statistical frame to assess the probability of observing a given tricluster in the given tensor data, accommodating principles to handle both tensor data and multivariate time series data with varying regularities. 
The methodology expands upon two state-of-the-art stances on the statistical significance of biclustering solutions: 1) \textsf{BSig} algorithm, proposed by \citep{henriques2018bsig}, where binomial testing is applied flexibly considering the underlying data regularities (e.g., either empirical or theoretical stances for real-valued and categorical data) and corrected for multiple hypotheses; 
and 2) \textsf{CCC-Biclustering} algorithm, proposed by \cite{madeira2008identification} for time series data, where patterns have a well-established duration and their probability is modeled under a first-order Markov chain, followed by a binomial significance test and correction for temporal misalignments \citep{gonccalves2010bimotif}.


To correctly calculate the probability of $ p_{\varphi_{T}}$, we must make assumptions regarding variable and context dependency (mutually dependent (MD), mutually independent (MI), identically distributed, non-identically distributed, or temporally contiguous (TC)). To this end, it is important to understand the type of data at hand. Table \ref{data_examples} comprises data of multiple domains, 
providing a view on which of the aforementioned assumptions can be placed to adequately create a null data model. 
For example, dependency between genes in omic data domains should be considered as diverse gene sets are strongly co-expressed in various pathways, affecting probability calculus \cite{chetty2009multiclass}. 
This type of analysis is crucial to correctly assess null statistical expectations as they determine the final significance estimates. 


\begin{table}[!t]
\caption{\footnotesize Null model assumptions for diverse real-world three-way data. Data is described according to \textit{observations-variables-context} axes and the underlying satisfied assumptions.}
\vskip -0.15cm
\begin{adjustwidth}{-2.2cm}{3cm}
\scriptsize
\begin{tabular}{l|c|c|c|c|c}
\toprule
\multicolumn{1}{c|}{data}                                             & \multicolumn{1}{c|}{\makecell{Variables MD \\ Context MD}} & \multicolumn{1}{c|}{\makecell{Variables MD \\ Context MI} $|$ \makecell{Variables MI \\ Context MD}} & \multicolumn{1}{c|}{\makecell{Variables MD \\ Context TC}} & \multicolumn{1}{c|}{\makecell{Variables MI \\ Context TC}} & \multicolumn{1}{c}{\makecell{Variables MI \\ Context MI}} \\ \midrule
\multicolumn{1}{l|}{\textit{station-day cluster-month}} & \multicolumn{1}{c|}{$\checkmark$}      & \multicolumn{1}{c|}{$\checkmark$}                 & \multicolumn{1}{c|}{}       & \multicolumn{1}{c|}{}       & \multicolumn{1}{c}{$\checkmark$}      \\ 
\multicolumn{1}{l|}{\textit{movie-keywords-celeb}}                    & \multicolumn{1}{c|}{$\checkmark$}      & \multicolumn{1}{c|}{$\checkmark$}                 & \multicolumn{1}{c|}{}       & \multicolumn{1}{c|}{}       & \multicolumn{1}{c}{$\checkmark$}      \\ 
\multicolumn{1}{l|}{\textit{gene-marker-tumor}}                       & \multicolumn{1}{c|}{$\checkmark$}       & \multicolumn{1}{c|}{$\checkmark$}                 & \multicolumn{1}{c|}{}       & \multicolumn{1}{c|}{}       & \multicolumn{1}{c}{$\checkmark$}      \\ 
\multicolumn{1}{l|}{\textit{subject-clustered pairs-time windows (fMRI)}}                   & \multicolumn{1}{c|}{$\checkmark$}      & \multicolumn{1}{c|}{$\checkmark$}                 & \multicolumn{1}{c|}{}       & \multicolumn{1}{c|}{}       & \multicolumn{1}{c}{$\checkmark$}      \\ 
\multicolumn{1}{l|}{\textit{job-jobseekers-seekerskills}}             & \multicolumn{1}{c|}{}      & \multicolumn{1}{c|}{$\checkmark$}                 & \multicolumn{1}{c|}{}       & \multicolumn{1}{c|}{}       & \multicolumn{1}{c}{$\checkmark$}      \\ 
\multicolumn{1}{l|}{\textit{individual-signal-signal}}                & \multicolumn{1}{c|}{}      & \multicolumn{1}{c|}{$\checkmark$}                 & \multicolumn{1}{c|}{}       & \multicolumn{1}{c|}{}       & \multicolumn{1}{c}{$\checkmark$}      \\ 
\multicolumn{1}{l|}{\textit{position-position-image/variable}}        & \multicolumn{1}{c|}{}       & \multicolumn{1}{c|}{$\checkmark$}                 & \multicolumn{1}{c|}{}       & \multicolumn{1}{c|}{}       & \multicolumn{1}{c}{$\checkmark$}      \\ 
\multicolumn{1}{l|}{\textit{movie-genre-tags}}                        & \multicolumn{1}{c|}{}       & \multicolumn{1}{c|}{$\checkmark$}                 & \multicolumn{1}{c|}{}       & \multicolumn{1}{c|}{}       & \multicolumn{1}{c}{$\checkmark$}      \\ 
\multicolumn{1}{l|}{\textit{user-behavior-environment}}               & \multicolumn{1}{c|}{}       & \multicolumn{1}{c|}{$\checkmark$}                 & \multicolumn{1}{c|}{}       & \multicolumn{1}{c|}{}       & \multicolumn{1}{c}{$\checkmark$}      \\ 
\multicolumn{1}{l|}{\textit{stations/provider-variables-time}}        & \multicolumn{1}{c|}{}       & \multicolumn{1}{c|}{}                  & \multicolumn{1}{c|}{$\checkmark$}      & \multicolumn{1}{c|}{$\checkmark$}      & \multicolumn{1}{c}{}       \\ 
\multicolumn{1}{l|}{\textit{sample-gene-time}}                             & \multicolumn{1}{c|}{}       & \multicolumn{1}{c|}{}                  & \multicolumn{1}{c|}{$\checkmark$}      & \multicolumn{1}{c|}{$\checkmark$}      & \multicolumn{1}{c}{}       \\ 
\multicolumn{1}{l|}{\textit{patient/provider-variable-time}}          & \multicolumn{1}{c|}{}       & \multicolumn{1}{c|}{}                  & \multicolumn{1}{c|}{$\checkmark$}      & \multicolumn{1}{c|}{$\checkmark$}      & \multicolumn{1}{c}{}       \\ 
\multicolumn{1}{l|}{\textit{vertex-vertex-time}}                      & \multicolumn{1}{c|}{}       & \multicolumn{1}{c|}{}                  & \multicolumn{1}{c|}{}       & \multicolumn{1}{c|}{$\checkmark$}      & \multicolumn{1}{c}{}       \\ 
\multicolumn{1}{l|}{\textit{users-bookmarks-tags (topics)}}           & \multicolumn{1}{c|}{}       & \multicolumn{1}{c|}{}                  & \multicolumn{1}{c|}{}       & \multicolumn{1}{c|}{}       & \multicolumn{1}{c}{$\checkmark$}      \\ \bottomrule
\multicolumn{6}{l}{} 
\label{data_examples}
\end{tabular}
\end{adjustwidth}
\vskip -0.35cm
\end{table}

\subsection{Tricluster pattern's probability}
Given a tricluster $T$ with pattern $\varphi_{T}$, if we assume that variables are mutually independent and that contexts are also mutually independent, then the probability of the tricluster's pattern occurring is given by

\begin{equation}
    \begin{aligned}
    p_{\varphi_{T}} = \prod_{z_k \in K}^{} p_{\varphi_{B_k}} = \prod_{z_k \in K}^{} \prod_{y_j \in J}^{} P(c_{jk}),
    \end{aligned}
\end{equation}

\noindent where $K$ comprises the contexts in the tricluster, $J$ is the set of variables in the tricluster, $P(c_{jk})$ is an estimate of the probability of value $c_{jk}$ occurrence, 
in the absence of adjustments and noise. As both variables and contexts are mutually independent, the probability of the value $c_{jk}$ is approximated by the available $n$ observations measured by variable $y_j$ in context $z_k$.

If we consider variables to be mutually dependent and contexts to be mutually independent, then a tricluster pattern's probability is given by

\begin{equation}
    p_{\varphi_{T}} = \prod_{z_k \in K}^{} p_{\varphi_{B_k}} = \prod_{z_k \in K}^{} P(\bigwedge\limits_{y_j \in J}^{} c_{jk}).
\end{equation}

\begin{figure}[!t]
\begin{adjustwidth}{0cm}{0cm}

\centering

\includegraphics[width=.98\textwidth]{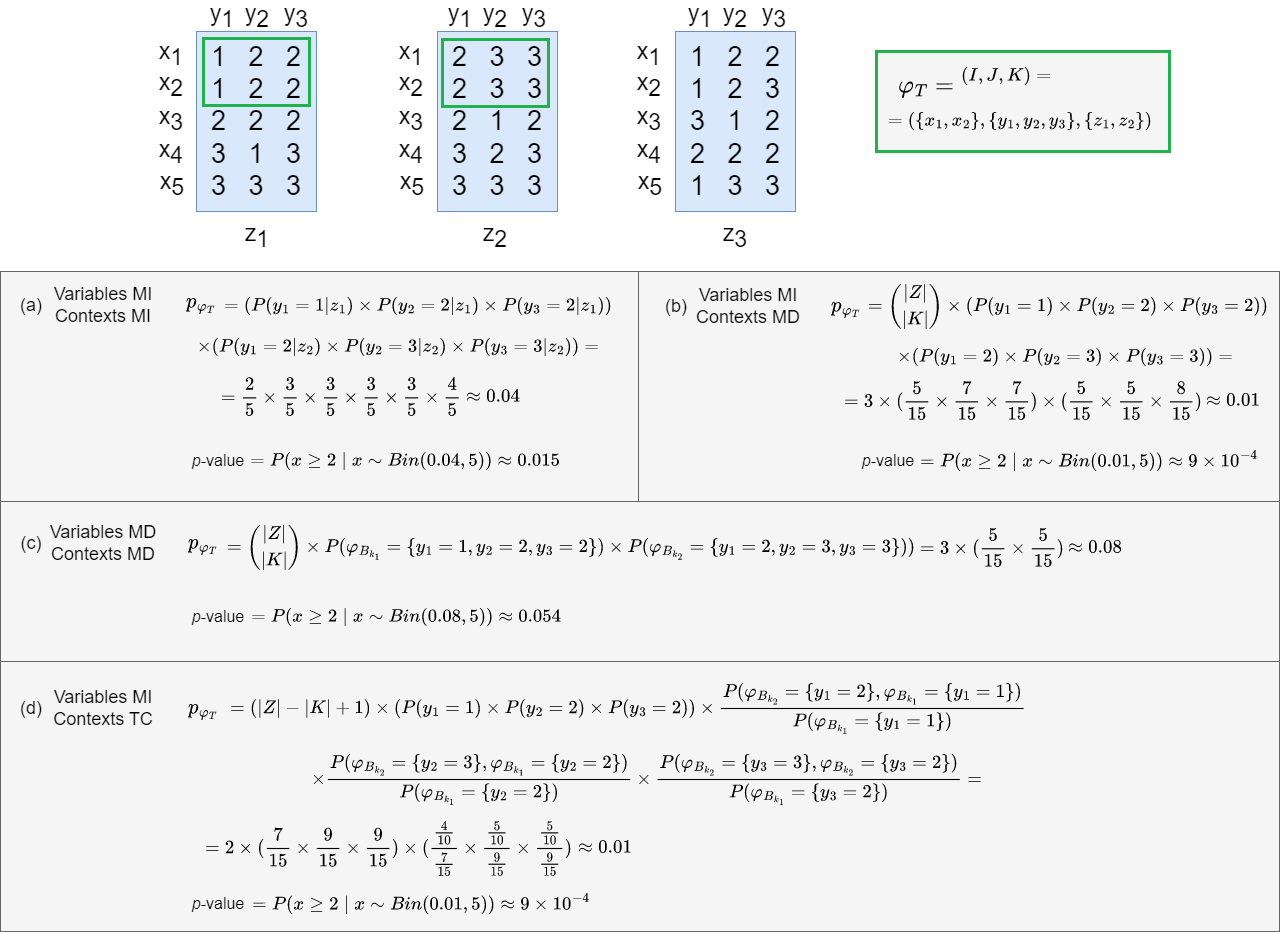}\\

\caption{\footnotesize Illustrative example of statistical significance calculus under different data assumptions.}
\label{probabilities_examples}
\end{adjustwidth}
\end{figure}

When contexts are mutually dependent, the probability of value $c_{jk}$ is approximated by the set of observations measured at $y_j$ across all contexts, $Z$. Additionally, as we need to account for the pattern to occur in any context of $Z$, 
then 

\begin{equation}
    \begin{aligned}
    p_{\varphi_{T}} = {|Z| \choose |K|}\prod_{z_k \in K}^{} p_{\varphi_{B_k}} = {|Z| \choose |K|}\prod_{z_k \in K}^{} \prod_{y_j \in J}^{} P(c_{jk}),
    \end{aligned}
\end{equation}

\noindent when variables are mutually independent and 

\begin{equation}
    p_{\varphi_{T}} = {|Z| \choose |K|}\prod_{z_k \in K}^{} p_{\varphi_{B_k}} = {|Z| \choose |K|}\prod_{z_k \in K}^{} P(\bigwedge\limits_{y_j \in J}^{} c_{jk}).
\end{equation}



\noindent when variables are mutually dependent. To further clarify the calculus, an example of the above formulas, using dummy data, is given in Figure \ref{probabilities_examples}.

In the presence of temporal data, where $Z$ represents the set of time points and $K$ is a contiguous subset of $Z$, then the null probability of a tricluster's pattern, $p_{\varphi_{T}}$, can be modeled by a first-order Markov chain. Instead of accounting for the possibility of the pattern occurring in any context of $Z$, we need to accommodate the possibility for temporal misalignments \citep{gonccalves2010bimotif}. To this end, $p_{\varphi_{T}}$ is given by

\begin{equation}
    p_{\varphi_{T}} =  (|Z| - |K| + 1) \times p_{\varphi_{B_{k_1}}} \times \prod_{z_k \in K \backslash z_{k_1}}^{} p_{\varphi_{B_k}|\varphi_{B_{k-1}}}
\end{equation}


\noindent where $p_{\varphi_{B_{k_1}}}$ is given by $\prod_{y_j \in J}^{} P(c_{jk_1})$ if variables are assumed to be mutually independent or $P(\bigwedge\limits_{y_j \in J}^{} c_{jk_1})$ if we assume mutual dependence, and $p_{\varphi_{B_k}|\varphi_{B_{k-1}}}$ represents the probability of the pattern transitioning from context $z_{k-1}$ to context $z_k$. The probability of transitioning depends on whether we assume variables to be mutually dependent or independent. If we assume variables to be mutually independent, then 

\begin{equation}
p_{\varphi_{B_k}|\varphi_{B_{k-1}}} = \prod_{y_j \in J}^{}\frac{P( c_{jk}, c_{j(k-1)})}{P(c_{j(k-1)})},
\end{equation}
\vskip 0.3cm

\noindent and, when assuming mutually dependent variables, then

\begin{equation}
p_{\varphi_{B_k}|\varphi_{B_{k-1}}} = \frac{ P(\bigwedge\limits_{y_j \in J}^{}c_{jk}, c_{j(k-1)})}{P(\bigwedge\limits_{y_j \in J}^{}c_{j(k-1)})}.
\end{equation}


\noindent Figure \ref{probabilities_examples} (d) presents an example of the aforesaid calculus when in the presence of mutually independent variables. 

The introduced probabilistic stances can be considered for categorical and real-valued variables. Empirical and theoretical distributions can be considered. The similarity between variable distributions can be assessed using 
$\chi^2$ goodness-of-fit test for categorical/ordinal data and Kolmogorov–Smirnov goodness-of-fit for real-valued data. For details on how to circumvent point estimate problems in real-valued data, we redirect the reader to 
\cite{henriques2018bsig}. 

\subsection{Tricluster's statistical significance}

Given $p_{\varphi_{T}}$, binomial tails can be used to robustly compute the probability of a tricluster $T = (I, J, K)$ with pattern $\varphi_{T}$ to occur for $|I|$ or more observations \citep{henriques2018bsig}. Accordingly, 

\begin{equation}
\text{\textit{p}-value} =  P(x\ge |I| \mid x\sim Bin(p_{\varphi_{T}}, |X|)) = \sum_{x=|I|}^{|X|} {|X| \choose x} p_{\varphi_{T}}^x (1-p_{\varphi_{T}})^{|X| - x},
\end{equation}

\noindent where $Bin(p_{\varphi_{T}},|X|)$ is the null binomial distribution, $p_{\varphi_{T}}$ the event probability, $|X|$ the sample size, and $|I|$ the number of successes.

When variables are identically distributed, the \textit{p}-value can be further corrected when considering that the observed pattern can span alternative variables \citep{henriques2018bsig}, i.e., 

\begin{equation}
    \text{\textit{p}-value} = {|Y|\choose |J|} \times P(x\ge |I| \mid x\sim Bin(p_{\varphi_{T}}, |X|))
\end{equation}


Irrespectively of the above relaxation, further corrections should be placed for multiple hypotheses (patterns). To this end, the Benjamini-Hochberg procedure \citep{benjamini1995controlling} is suggested to decrease the false discovery rate. This correction 
tackles the problems faced by more conservative corrections, such as Bonferroni, in the presence of a high number of hypotheses.


Instantiating the introduced methodology, Table \ref{min_n} shows the minimum number of observations required for a tricluster's pattern $\varphi_T$ to be statistically significant ($\alpha < 0.01$). 
In particular, given a three-way data with uniformly distributed variables, 
this analysis shows how the minimum number of observations varies for: i) patterns with varying number of variables $|J|$ and contexts $|K|$, ii) variables with varying cardinality ($|L|$), iii) data with varying dependency assumptions, and iv) the presence of multi-hypotheses corrections. 
We confirm that ``Variables MI, Contexts MD" assumption is more conservative than ``Variables MI, Contexts TC", 
generally requiring fewer tricluster's observations to 
guarantee statistical significance. 
The pattern size ($|J|$ and $|K|$) heavily influence significance, with $|K|$ being particularly relevant under the latter assumption. 

\begin{adjustwidth}{-2cm}{.5cm}
\begin{minipage}{1.25\textwidth}

\captionof{table}{\footnotesize Given uniformly distributed three-way data ($|X|$=1000, $|Y|$=50, $|Z|$=50), the minimum number of observations of a tricluster, $n_{min}$, required to be statistically significant (\textit{p}-value $<$ 0.01) is assessed. In particular, we measure the impact of pattern size (number of variables $|J|$ and contexts, $|K|$), cardinality ($|L|$), and corrections. Two major assumptions are considered to this end, ``Variables MI, Contexts MD" where P($\varphi_{T}$) = $\frac{1}{|L|}^{|J| \times |K|} \times {|Z| \choose |K|}$ and ``Variables MI, Contexts TC" where P($\varphi_{T}$) = $\frac{1}{|L|}^{|J| \times |K|} \times (|Z| - |K| + 1)$.}
\scriptsize
\begin{tabular}{l|l|llll|llll}
\multicolumn{10}{c}{\textbf{$|X| = 1000, |Y| = 50, |Z| = 50$} (variables with Uniform distribution)}\\\midrule[1.5pt]
\multicolumn{2}{c}{}& \multicolumn{4}{c}{Variables MI, Contexts MD}                                                      & \multicolumn{4}{|c}{Variables MI, Contexts TC}                                                                  \\ \midrule
                     \multicolumn{1}{c}{} & \multicolumn{1}{c|}{$|J|$}                               & 2                    & 2                    & 2                    & 2                     & 2                    & 2                    & 2                     & 2                     \\
                     \multicolumn{1}{c}{} & \multicolumn{1}{c|}{$|K|$}                               & 2                    & 3                    & 4                    & 5                     & 2                    & 3                    & 4                     & 5                     \\ \midrule
\multirow{3}{*}{$|L|$=3} & $n_{min}$                       & -                 & -                 & -                 & -                  & 641                  & 85                   & 14                    & 3                     \\
                     & $n_{min}$  (corrected) & -                 & -                 & -                 & -                  & 671                  & 102                  & 21                    & 7                     \\
                     & P($\varphi_{T}$)                  & 15.12                    & 26.88                    & 35.10                    & 35.88                     & 0.60                 & 0.07                 & $7.16$$\times$$10^{-3}$  & $7.79$$\times$$10^{-4}$  \\ \midrule
\multirow{3}{*}{$|L|$=5} & $n_{min}$                       & -                 & -                 & 626                  & 248                   & 99                   & 8                    & 1                     & 1                     \\
                     & $n_{min}$  (corrected) & -                 & -                 & 656                  & 275                   & 117                  & 13                   & 3                     & 2                     \\
                     & P($\varphi_{T}$)                   & 1.96                    &  1.25                    & 0.59                 & 0.22                  & 0.08                 & $3.07$$\times$$10^{-3}$ & $1.20$$\times$$10^{-4}$  & $4.71$$\times$$10^{-6}$  \\ \midrule[1.5pt]
                     \multicolumn{1}{c}{} & \multicolumn{1}{c|}{$|J|$}                               & 3                    & 3                    & 3                    & 3                     & 3                    & 3                    & 3                     & 3                     \\
                     \multicolumn{1}{c}{} & \multicolumn{1}{c|}{$|K|$}                               & 2                    & 3                    & 4                    & 5                     & 2                    & 3                    & 4                     & 5                     \\ \midrule
\multirow{3}{*}{$|L|$=3} & $n_{min}$                       & -                 & -                 & 470                  & 174                   & 86                   & 7                    & 1                     & 1                     \\
                     & $n_{min}$  (corrected) & -                 & -                 & 510                  & 205                   & 109                  & 13                   & 4                     & 2                     \\
                     & P($\varphi_{T}$)                   & 1.68                    & 0.99                 & 0.43                 & 0.14                  & 0.07                 & $2.43$$\times$$10^{-3}$ & $8.84$$\times$$10^{-5}$  & $3.20$$\times$$10^{-6}$  \\ \midrule
\multirow{3}{*}{$|L|$=5} & $n_{min}$                       & 99                   & 18                   & 4                    & 1                     & 8                    & 1                    & 1                     & 1                     \\
                     & $n_{min}$  (corrected) & 123                  & 29                   & 9                    & 4                     & 15                   & 3                    & 1                     & 1                     \\
                     & P($\varphi_{T}$)                   & 0.08                 & 0.01                 & $9.43$$\times$$10^{-4}$ & $6.94$$\times$$10^{-5}$  & $3.13$$\times$$10^{-3}$ & $2.45$$\times$$10^{-5}$ & $1.92$$\times$$10^{-7}$  & $1.51$$\times$$10^{-9}$  \\ \midrule[1.5pt]
                     \multicolumn{1}{c}{} & \multicolumn{1}{c|}{$|J|$}                               & 4                    & 4                    & 4                    & 4                     & 4                    & 4                    & 4                     & 4                     \\
                     \multicolumn{1}{c}{} & \multicolumn{1}{c|}{$|K|$}                               & 2                    & 3                    & 4                    & 5                     & 2                    & 3                    & 4                     & 5                     \\ \midrule
\multirow{3}{*}{$|L|$=3} & $n_{min}$                       & 216                  & 51                   & 11                   & 3                     & 14                   & 1                    & 1                     & 1                     \\
                     & $n_{min}$  (corrected) & 255                  & 73                   & 22                   & 8                     & 26                   & 5                    & 2                     & 1                     \\
                     & P($\varphi_{T}$)                   & 0.18                 & 0.04                 & $5.35$$\times$$10^{-3}$ & $6.07$$\times$$10^{-4}$  & $7.46$$\times$$10^{-3}$ & $9.03$$\times$$10^{-5}$ & $1.09$$\times$$10^{-6}$  & $1.31$$\times$$10^{-8}$  \\ \midrule
\multirow{3}{*}{$|L|$=5} & $n_{min}$                       & 8                    & 1                    & 1                    & 1                     & 1                    & 1                    & 1                     & 1                     \\
                     & $n_{min}$  (corrected) & 16                   & 4                    & 1                    & 1                     & 5                    & 1                    & 1                     & 1                     \\
                     & P($\varphi_{T}$)                   & $3.14$$\times$$10^{-3}$ & $8.03$$\times$$10^{-5}$ & $1.51$$\times$$10^{-6}$ & $2.22$$\times$$10^{-6}$  & $1.25$$\times$$10^{-4}$ & $1.97$$\times$$10^{-7}$ & $3.08$$\times$$10^{-10}$ & $4.82$$\times$$10^{-13}$ \\ \midrule[1.5pt]
                     \multicolumn{1}{c}{} & \multicolumn{1}{c|}{$|J|$}                               & 5                    & 5                    & 5                    & 5                     & 5                    & 5                    & 5                     & 5                     \\
                     \multicolumn{1}{c}{} & \multicolumn{1}{c|}{$|K|$}                               & 2                    & 3                    & 4                    & 5                     & 2                    & 3                    & 4                     & 5                     \\ \midrule
\multirow{3}{*}{$|L|$=3} & $n_{min}$                       & 32                   & 5                    & 1                    & 1                     & 4                    & 1                    & 1                     & 1                     \\
                     & $n_{min}$  (corrected) & 51                   & 12                   & 5                    & 2                     & 10                   & 3                    & 1                     & 1                     \\
                     & P($\varphi_{T}$)                   & 0.20                 & $1.36$$\times$$10^{-3}$ & $6.60$$\times$$10^{-5}$ & $2.50$$\times$$10^{-6}$  & $8.29$$\times$$10^{-4}$ & $3.34$$\times$$10^{-6}$ & $1.34$$\times$$10^{-8}$  & $5.42$$\times$$10^{-11}$ \\ \midrule
\multirow{3}{*}{$|L|$=5} & $n_{min}$                       & 1                    & 1                    & 1                    & 1                     & 1                    & 1                    & 1                     & 1                     \\
                     & $n_{min}$  (corrected) & 6                    & 2                    & 1                    & 1                     & 3                    & 1                    & 1                     & 1                     \\
                     & P($\varphi_{T}$)                   & $1.25$$\times$$10^{-4}$ & $6.42$$\times$$10^{-7}$ & $2.41$$\times$$10^{-9}$ & $7.11$$\times$$10^{-12}$ & $5.02$$\times$$10^{-6}$ & $1.57$$\times$$10^{-9}$ & $4.92$$\times$$10^{-13}$ & $1.54$$\times$$10^{-16}$ \\ \bottomrule
                      
\end{tabular}
\label{min_n}
\end{minipage}
\end{adjustwidth}





\section{Computational and Experimental setup}

\paragraph{Triclustering Algorithms}
Triclustering algorithms 
can be categorized according to specific characteristics, including the underlying search procedure, targeted homogeneity, allowed data types (e.g., real-valued, ordinal, binary tensor data), amongst others \citep{henriques2018triclustering}. To fully explore the proposed solution, we selected three triclustering algorithms representative of three major search criteria -- heuristic-based (greedy), exhaustive based on formal concept analysis, and quasi-exhaustive based on biclustering-based consensus. Accordingly, the selected algorithms are 1) $\delta$-Trimax \citep{bhar2015multiobjective}, a greedy iterative search algorithm that handles real-valued tensor; 2) TRIAS algorithm \citep{jaschke2006trias}, an exhaustive algorithm that only handles binary data; and 3) modified version of Zaki \textit{TriCluster} algorithm \citep{soares2021towards}, a consensus-based algorithm with a modification that restricts the patterns to be contiguous along the context axis.

\paragraph{Datasets}
To validate the proposed methodology, three publically available \textbf{case studies} are selected: \textit{glycine} data \citep{irwin2016contribution}, \textit{gene expression} profile \citep{kanno2006per}, and \textit{penicillin batch fermentation} \citep{goldrick2015development, goldrick2019modern}. Table \ref{data_description} provides a brief description of each dataset and the applied preprocessing. 


\begin{table}[!b]
\caption{Summary of the selected case studies, including the size and domain of each dimension (e.g., batches, units, genes), and the applied preprocessing.}
\begin{adjustwidth}{-1cm}{0cm}
\scriptsize
\begin{tabular}{p{3.6cm}p{9.8cm}p{2.3cm}}\toprule
Datasets & Description & Preprocessing \\ \midrule
\textit{glycine} \mbox{(subjects x units x time)} \mbox{(23 x 467 x 6)} & A metabolomics dataset from urinary samples analyzed  by nuclear magnetic resonance spectroscopy giving  a total of 467 integrated units per NMR spectrum  for each observation. In this study, they used commercial  benzoic acid containing flavored water as a vehicle for  designed interventions. & \mbox{-- $|L|$=5} \mbox{(ordinal categories)} \\\midrule
\textit{gene expression} \newline \mbox{(samples x genes x time)} \mbox{(15 x 500 x 4)} & Liver tissue transcriptomics (45101 genes) from  male mice in response to the application of  tetrachlorodibenzodioxin (TCDD) administered  orally at doses of 0, 1, 3, 10 and 30$\mu g/kg$.  The liver was sampled 2, 4, 8 and 24 h after  administration. & \mbox{-- selection of top 500} \newline \mbox{genes with highest} \newline \mbox{variability on $X$$\times$$Z$;}\newline \mbox{-- $|L|$ = 3} \\\midrule
\mbox{\textit{penicillin batch fermentation}} \newline \mbox{(batches x variables x time)} \mbox{(100 x 16 x 25)} & An advanced mathematical model of a  100,000 litre penicillin fermentation system (bioprocess data).  It contains 100 batches, where batches 1-30 are  controlled by recipe-driven approach, 31-60 are  controlled by operators, 61-90 are controlled by  an Advanced Process Control (APC) solution using  the Raman spectroscopy, and 91-100 contain faults  resulting in process deviations. Each batch is a  multivariate time series where all available  processes, model inputs, and Raman spectroscopy  measurements are available. There isn't a fixed  number of time points per batch.  & \mbox{-- Removed Raman}\newline spectroscopy;\newline \mbox{-- Removed uninfor-} \mbox{mative variables;}\newline \mbox{-- Piecewise Aggregate} \newline \mbox{Approximation (PAA);} \newline \mbox{(25 time points)} \newline -- $|L| = 7$\\\bottomrule
\end{tabular}
\label{data_description}
\end{adjustwidth}
\end{table}

To complement the analysis, we generated \textbf{synthetic data} using G-Tric, a synthetic data generator proposed by 
\cite{lobo2021g}, able to generate tensors with planted triclusters. The authors provide a set of dataset sizes based on the type of data such as gene expression, financial, fMRI, social, and georeferenced time-series. With this in mind, the datasets created contained 1000 observations, 50 variables, 50 contexts, and two types of background distribution -- uniform and Gaussian ($\mu = 0.0, =30$) --, and 5 triclusters planted per dataset. The artificial variables are ordinal with cardinality $|L| \in \{3, 5, 7\}$. The planted triclusters had the following properties: 1) the number of observations between 50 and 500, 2) the number of variables between 2 and 4, 3) the number of contexts between 2 and 4, and 4) patterns could be contiguous or non-contiguous across contexts.

\paragraph{Pattern Evaluation}
In the synthetic data generated, variables hold no particular type of information, there is no prior knowledge about their dependency, and all of the variables are sampled from either a gaussian or uniform distribution. To this end, for the purpose of testing the proposed statistical frame, we assume variables to be identically distributed and the following cases regarding variable dependency: 1) variables to be mutually independent, and 2) variables to be mutually dependent. Regarding context dependency, we assume the following cases: 1) contexts to be mutually independent, 2) contexts to be mutually dependent, and 3) contexts to represent a set of time points.

Regarding the selected case studies, all of them contain one temporal dimension. 
This means that patterns that exhibit contiguity across this dimension can be evaluated assuming a first-order Markov chain. Out of the three algorithms, only the modified Zaki algorithm \citep{soares2021towards} extracts patterns contiguous across time. As such, the patterns extracted using the modified Zaki algorithm will be analyzed assuming a first-order Markov chain ,whilst the patterns extracted using TRIAS and $\delta$-Trimax will be analyzed assuming contexts to be mutually dependent. The variables dimension in the \textit{penicillin batch fermentation} and \textit{glycine} cases will be analyzed assuming variable independence, while the \textit{gene expression} case will be analyzed assuming variable mutual dependency. 


\section{Results and Discussion}

In order to assess the validity and implication of the proposed statistical frame, we tested the concepts of the proposed methodology in two major subsections: 1) case studies, and 2) synthetic data. 
The results acquired per case study provide an in-depth examination of statistically significant patterns. 
Synthetic data results allow us to assess expectations on what makes a tricluster to yield statistical significance for different data characteristics, as well as the impact of the placed corrections. 

\subsection*{Real-world case studies}

\begin{figure}
\begin{adjustwidth}{-1.8cm}{-1.8cm}
\vskip -0.5cm
\centering
\scriptsize
\textbf{I. \textit{Glycine} dataset} \vskip -0.25cm

\includegraphics[height=2.3in]{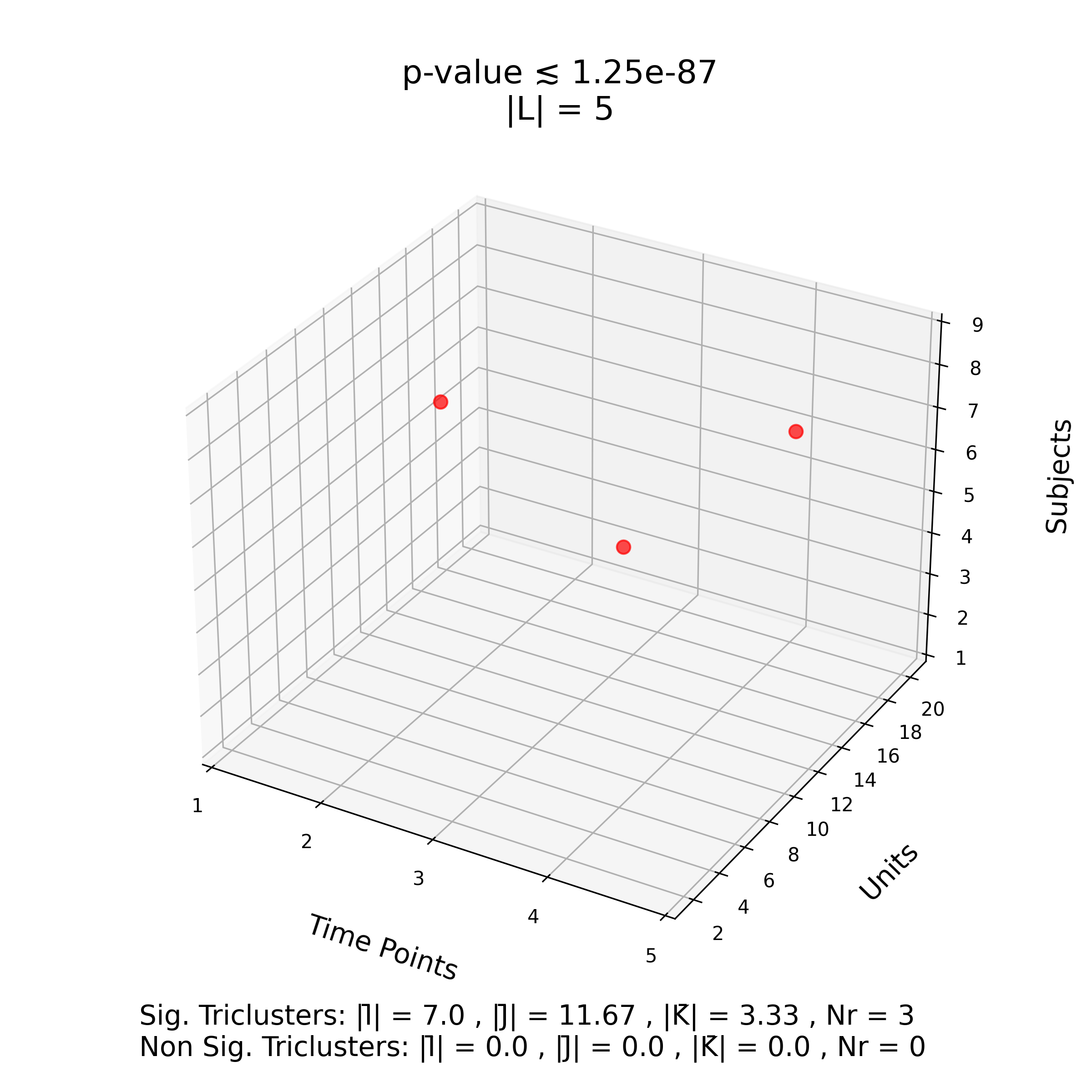}
\includegraphics[height=2.3in]{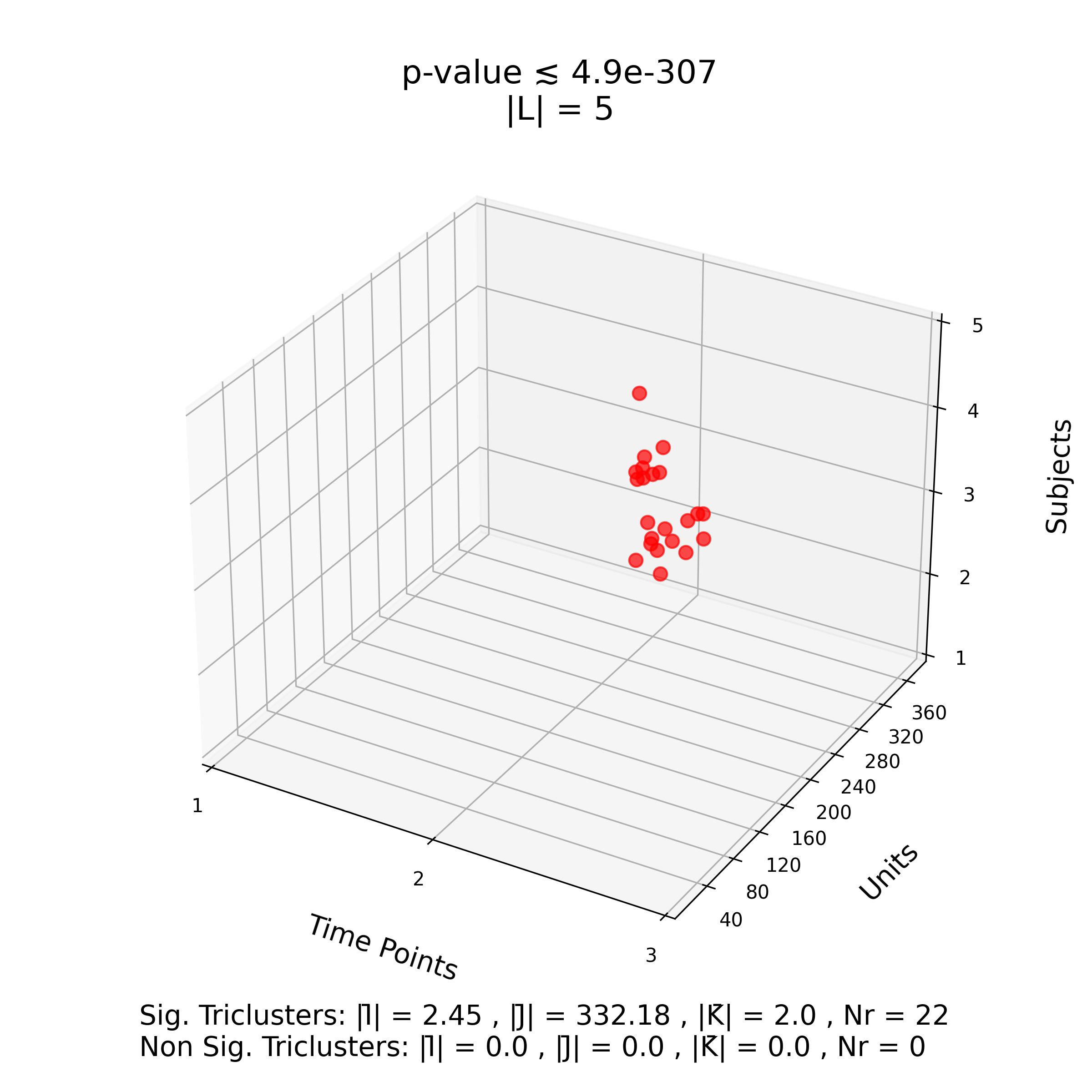}
\includegraphics[height=2.3in]{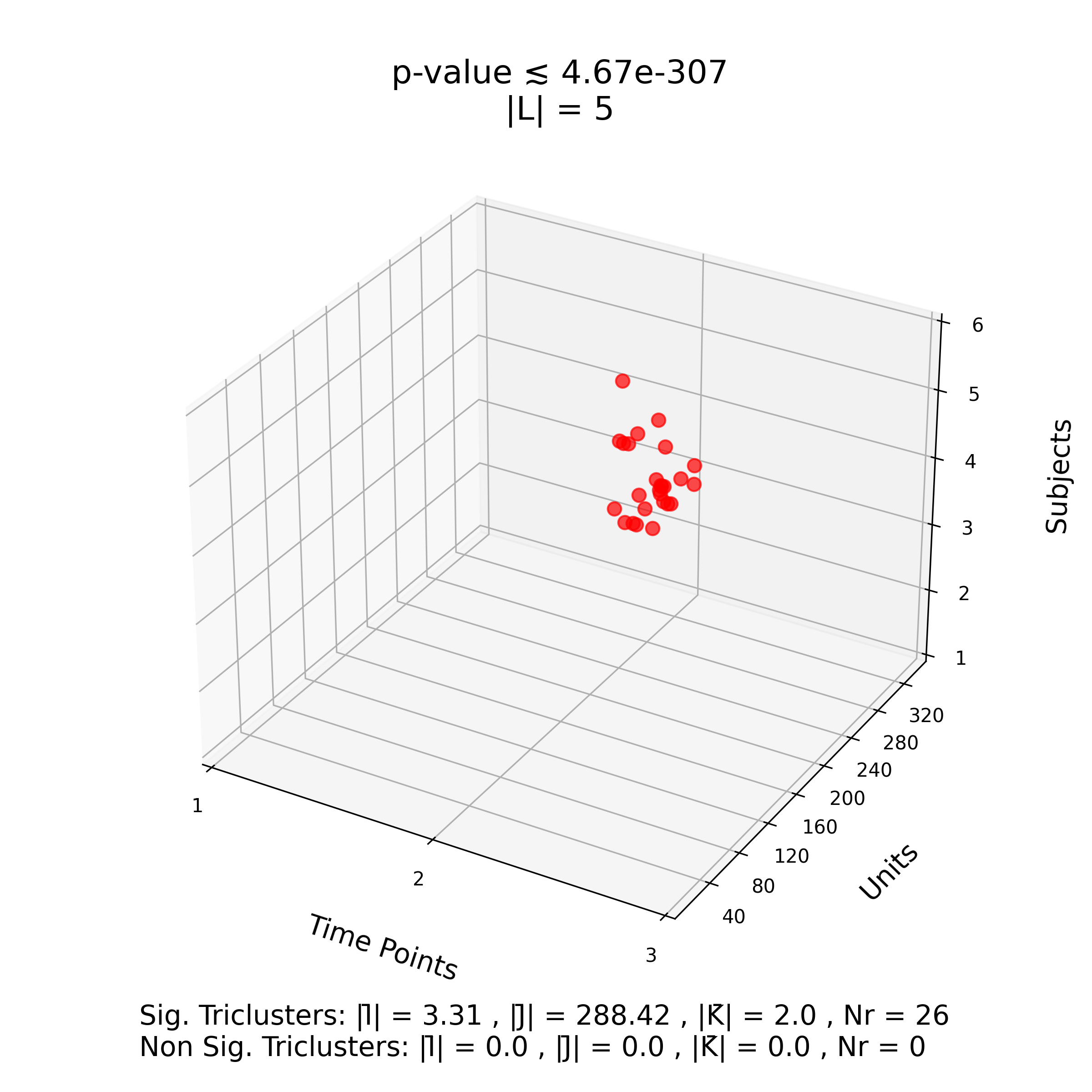} \\
\textbf{a)} $\delta$-Trimax \ \ \ \ \ \ \ \ \ \ \ \ \ \ \ \ \ \ \ \ \ \ \ \ \ \ \ \ \ \ \ \ \ \ \ \ \ \ \textbf{b)} \textit{TriCluster} \ \ \ \ \ \ \ \ \ \ \ \ \ \ \ \ \ \ \ \ \ \ \ \ \ \ \ \ \ \ \ \ \ \ \ \ \ \  \textbf{c)} TRIAS \\
\vskip 0.25cm
\textbf{II. \textit{Gene expression} dataset}\vskip -0.25cm

\includegraphics[height=2.3in]{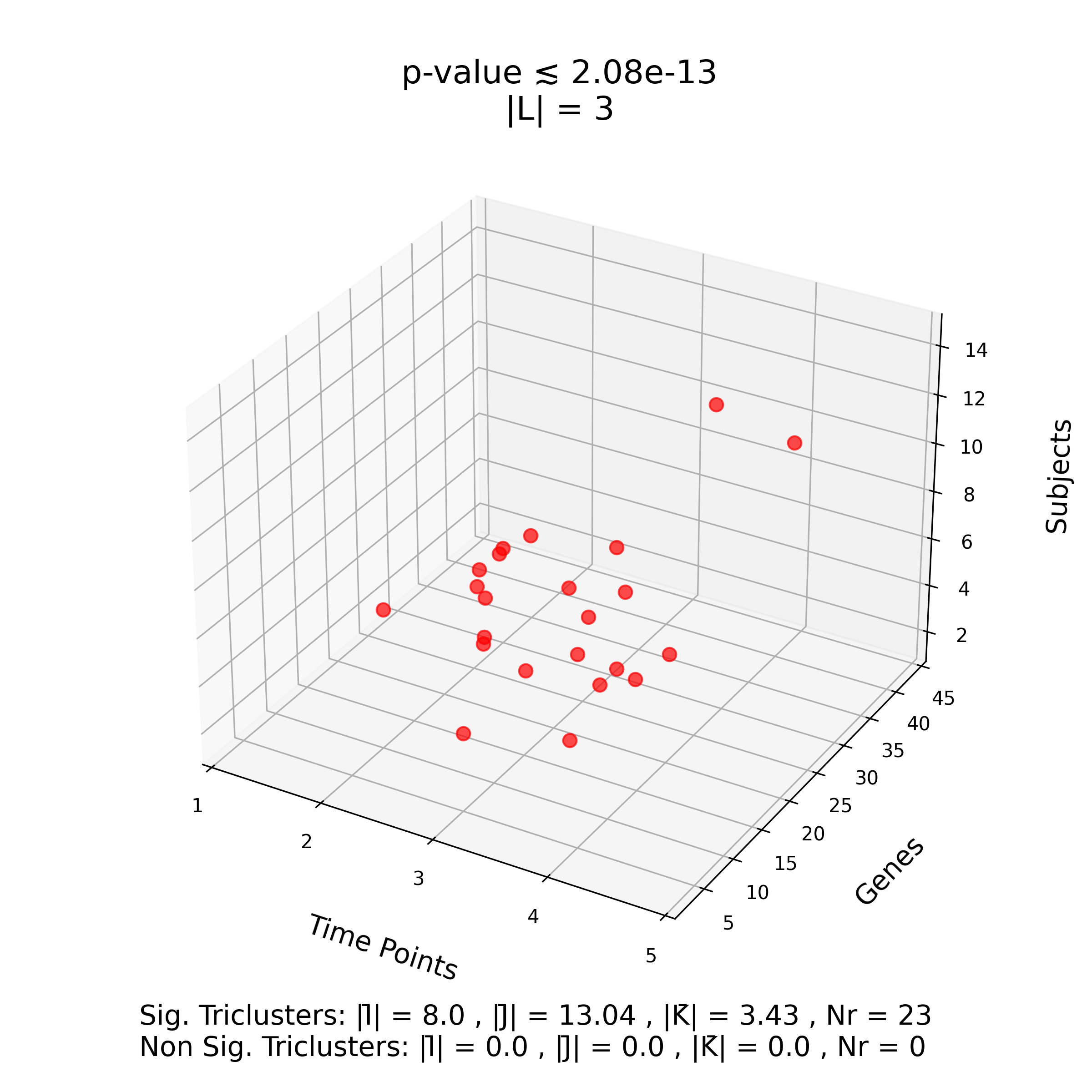}
\includegraphics[height=2.3in]{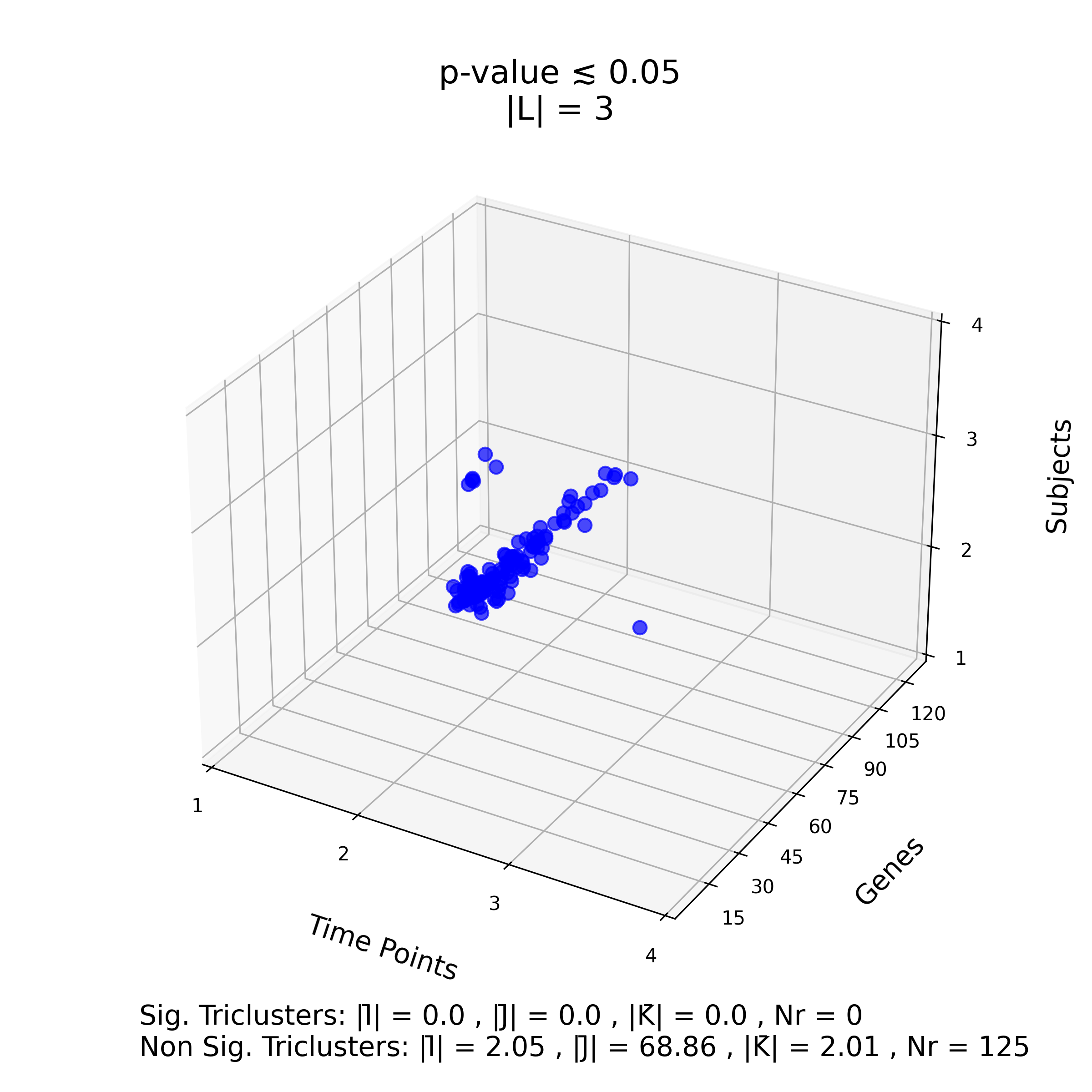}
\includegraphics[height=2.3in]{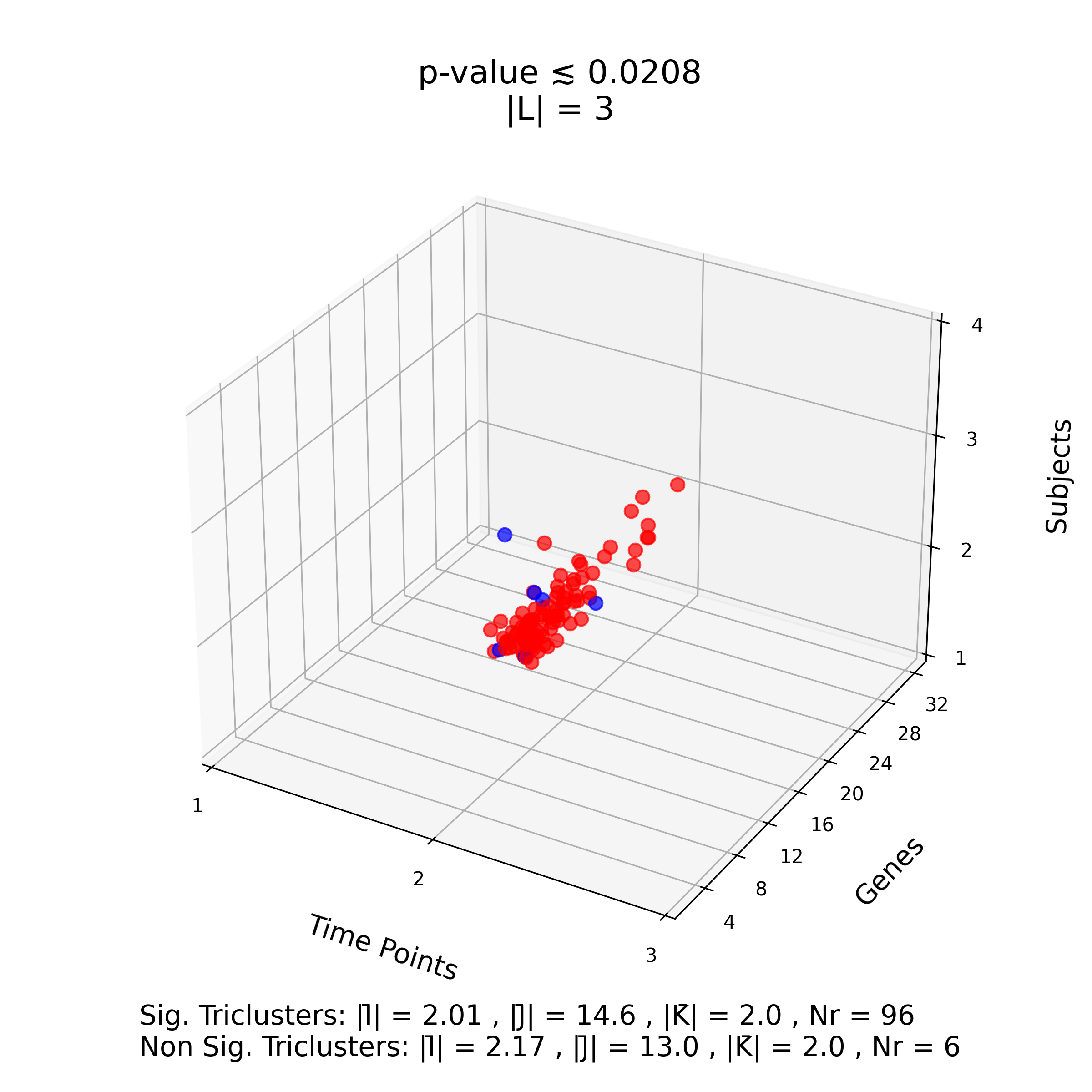} \\
\textbf{a)} $\delta$-Trimax \ \ \ \ \ \ \ \ \ \ \ \ \ \ \ \ \ \ \ \ \ \ \ \ \ \ \ \ \ \ \ \ \ \ \ \ \ \ \textbf{b)} \textit{TriCluster} \ \ \ \ \ \ \ \ \ \ \ \ \ \ \ \ \ \ \ \ \ \ \ \ \ \ \ \ \ \ \ \ \ \ \ \ \ \  \textbf{c)} TRIAS \\
\vskip 0.25cm
\textbf{III. \textit{Penicillin batch fermentation} dataset}\vskip -0.25cm

\includegraphics[height=2.3in]{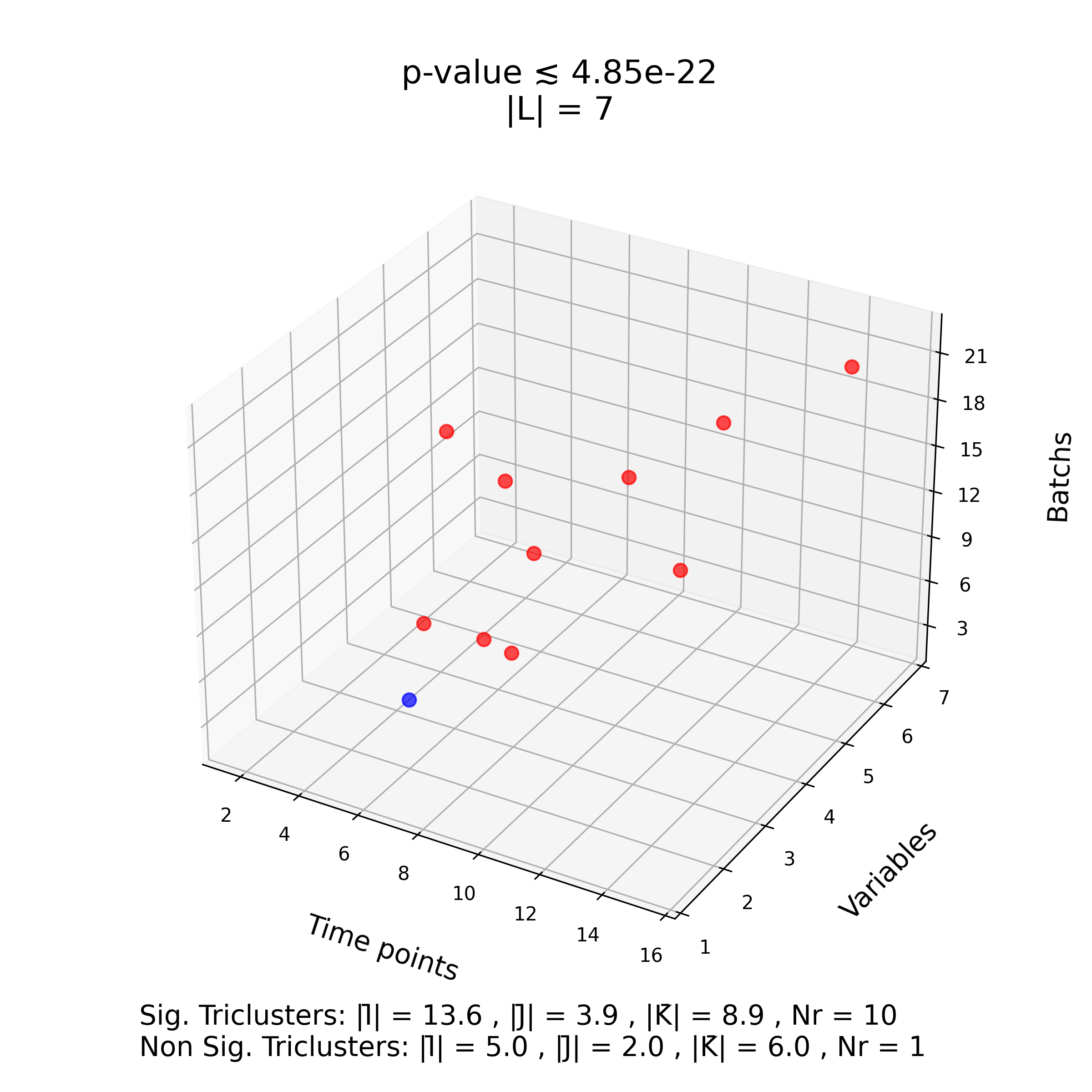}
\includegraphics[height=2.3in]{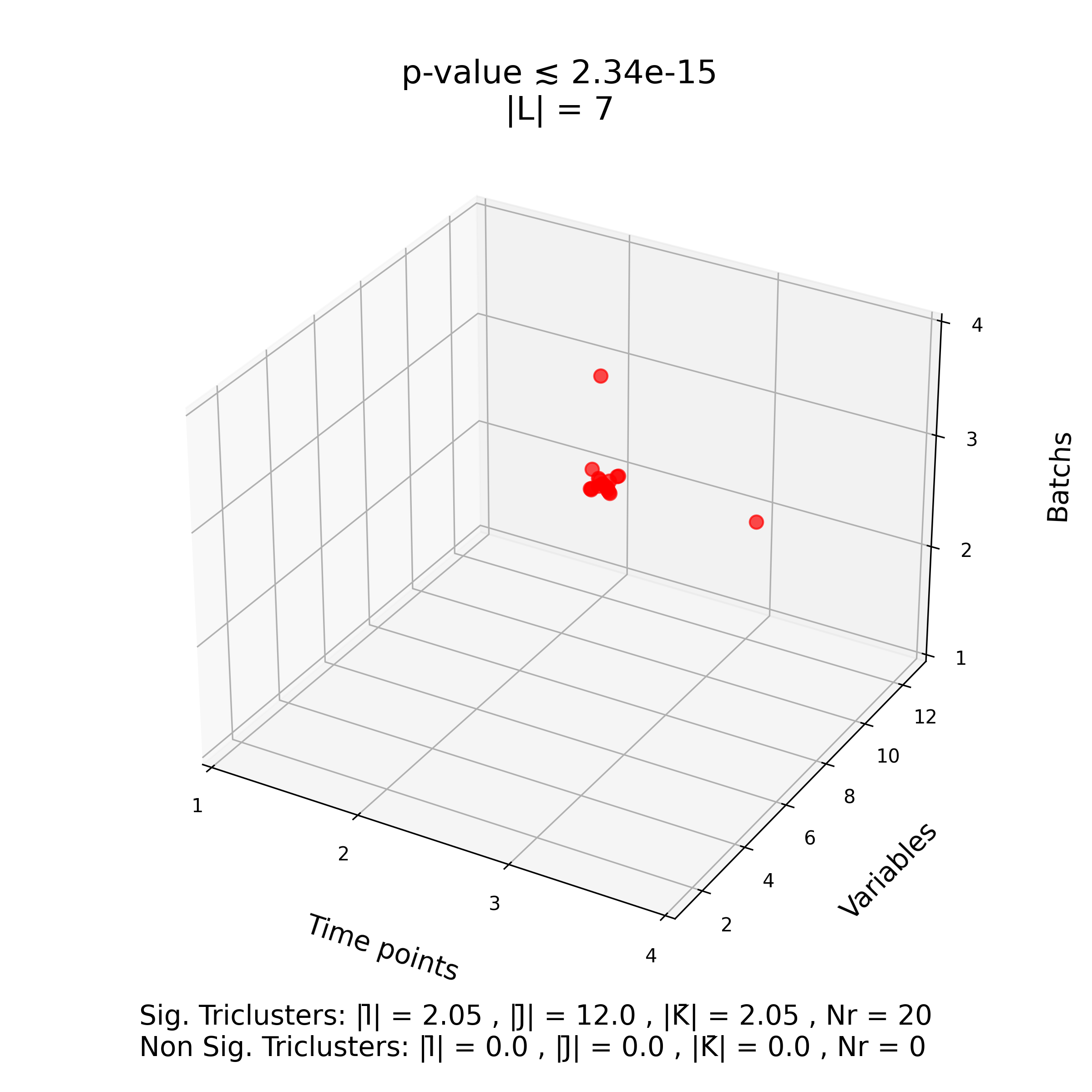}
\includegraphics[height=2.3in]{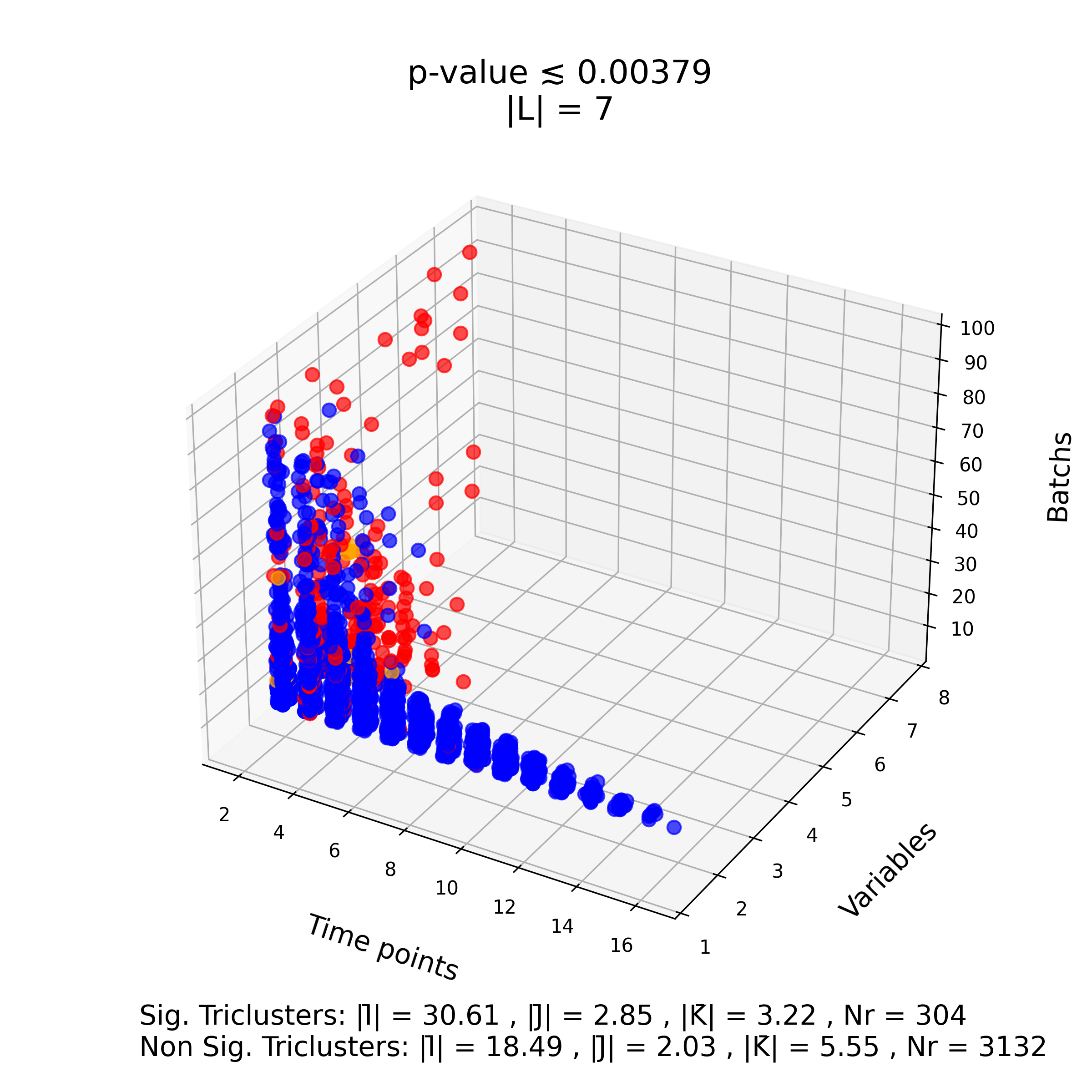} \\
\textbf{a)} $\delta$-Trimax \ \ \ \ \ \ \ \ \ \ \ \ \ \ \ \ \ \ \ \ \ \ \ \ \ \ \ \ \ \ \ \ \ \ \ \ \ \ \textbf{b)} \textit{TriCluster} \ \ \ \ \ \ \ \ \ \ \ \ \ \ \ \ \ \ \ \ \ \ \ \ \ \ \ \ \ \ \ \ \ \ \ \ \ \  \textbf{c)} TRIAS \\

\caption{\small Solutions $\Bar{I}$ returned from state-of-the-art triclustering algorithms ($\delta$-Trimax, \textit{TriCluster} and TRIAS) applied to three real-world datasets. Each dot on the scatter plot represents a tricluster where its position is defined by the number of observations ($|I|$), variables ($|J|$), and time points ($|K|$). The dots are colored red if they are below a statistical significance threshold defined by the Benjamini-Hochberg procedure, colored orange if they are above the threshold but below 0.05, and colored blue otherwise.}
\label{use_cases_glycine}
\end{adjustwidth}
\end{figure}

Consider the \textit{glycine} dataset, with results presented in Figure \ref{use_cases_glycine}-I. All algorithms extracted a small number of patterns, up to 26 patterns. The $\delta$-Trimax extracted a total of 3 patterns, with an average number of subjects of around 3 (30\%), an average number of 11 units (2\%), and an average of 3 time points (50\%). \textit{TriCluster} and TRIAS extracted slightly more patterns, 22 and 26 accordingly, but with a lower average of subjects, between 2 and 5 subjects (5\%-20\%), and a lower average of time points, only 2 time points in all of the patterns (33\%). Despite different assumptions considered, the patterns extracted by all of the algorithms generally yield statistical significance (low $p$-values). The Benjamini-Hochberg procedure sets the statistical significance threshold between $4.9 \times 10^{-307}$ and $1.25 \times 10^{-87}$ decimals. This can be explained by the imposed assumption in the dimension of the units, associated with extremely low co-occurrence probabilities. 

Consider the results of drug response gene expression presented in Figure \ref{use_cases_glycine}-II. In this case, $\delta$-Trimax extracted a total of 23 patterns, while \textit{TriCluster} and TRIAS extracted a total of 125 and 102 patterns respectively. Patterns extracted by $\delta$-Trimax have an average number of 8 subjects (67\%), 13 genes (3\%), and 3 time points (75\%). \textit{TriCluster} and TRIAS have a lower number of subjects (16\%), 68 genes (13\%) in \textit{TriCluster} and 14 genes (3\%) in TRIAS, and 2 time points (50\%). Contrary to the glycine case study, due to a more conservative null model assumption, all patterns extracted by \textit{TriCluster} have a \textit{p}-value above the imposed significance threshold. Similarly, some patterns extracted by TRIAS also fail to meet the imposed significance threshold. 

Results of the industrial batches' study are presented in Figure \ref{use_cases_glycine}-III. Both $\delta$-Trimax and the \textit{TriCluster} extracted a small number of patterns (11 and 20), whilst TRIAS retrieved over 3400 patterns. For the \textit{TriCluster} algorithm, all patterns yield statistical significance. The patterns extracted by \textit{TriCluster} occur on average in 2 batches (2\%), on average in 12 variables (75\%), and on average in 2 time points (8\%), meaning these patterns despite occurring on a very low number of batches have enough variables and time points to be marked as statistically significant. In the patterns extracted by $\delta$-Trimax we observe a threshold between significant and non-significant patterns, between 2 and 3 variables. Patterns with statistical significance contain on average 14 batches (13\%),  4 variables (25\%) and 9 time points (36\%). Contrary to previous case studies, TRIAS extracted 9\% of patterns with statistical significance and 3132 non-significant patterns (91\%). As depicted in Figure \ref{use_cases_glycine}-III, most non-significant patterns occur in patterns with a lower number of variables between 2 and 3 when compared against statistically significant patterns. In this case, we can observe that the number of time points does not condition the pattern's statistical significance even though some patterns occur up to 16 time points without statistical significance. 

\subsection*{Synthetic data}

What makes a tricluster statistically significant? To answer this research question, synthetic data with planted triclustering solutions are used to statistically assess triclusters with varying properties.

\begin{figure}[!t]
\begin{adjustwidth}{-1cm}{0cm}
\vspace*{-1cm}
\begin{center}
    \small Synthetic data with Uniform distribution
\end{center}

\begin{center}
\includegraphics[height=2.5in]{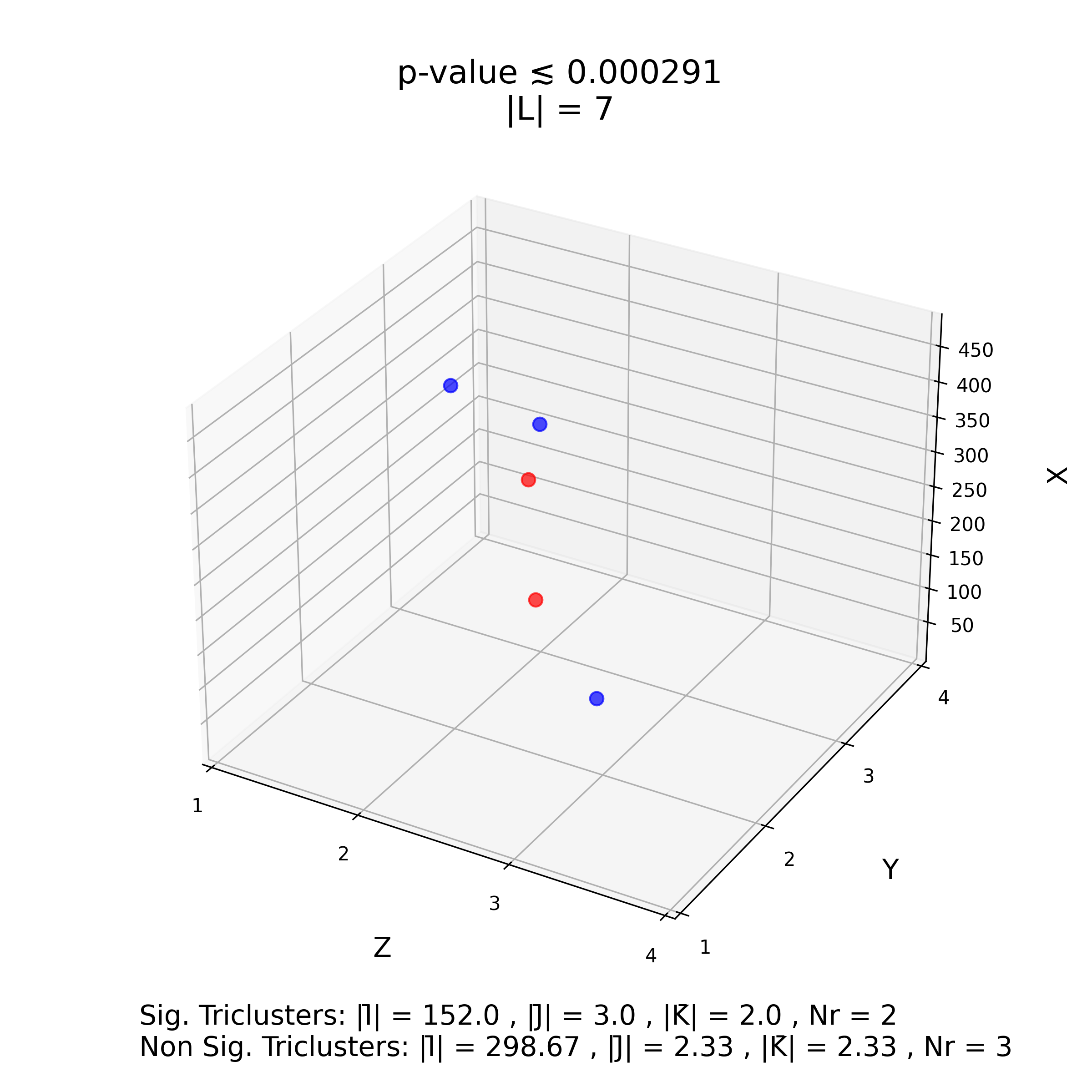}
\includegraphics[height=2.5in]{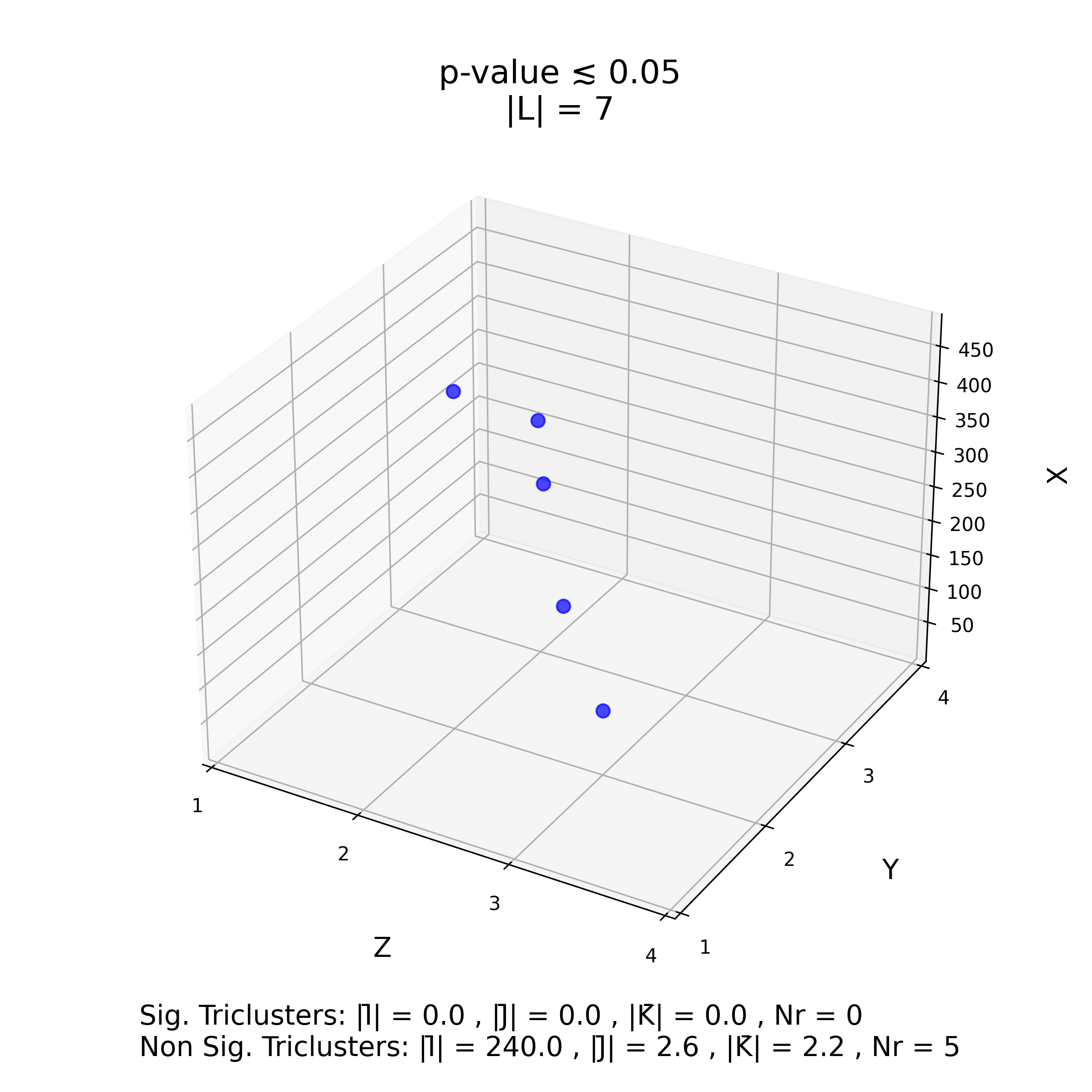}\\
\scriptsize{Without \textit{p}-value correction: \ \ \ \ \ \ \ \ \ \ \ \ \ \ \ \ \ \ \ \ \ \ \ \ With \textit{p}-value correction:} \\
\scriptsize{Variables Mutually dependent \ \ \ \ \ \ \ \ \ \ \ \ \ \ \ \ \ \ \ Variables Mutually dependent} \\
\scriptsize{Contexts Mutually dependent \ \ \ \ \ \ \ \ \ \ \ \ \ \ \ \ \ \ \ Contexts Mutually dependent} \\
\end{center}

\vspace{0.2cm}
\begin{center}
    \small Synthetic data with Gaussian distribution
\end{center}

\begin{center}
\includegraphics[height=2.5in]{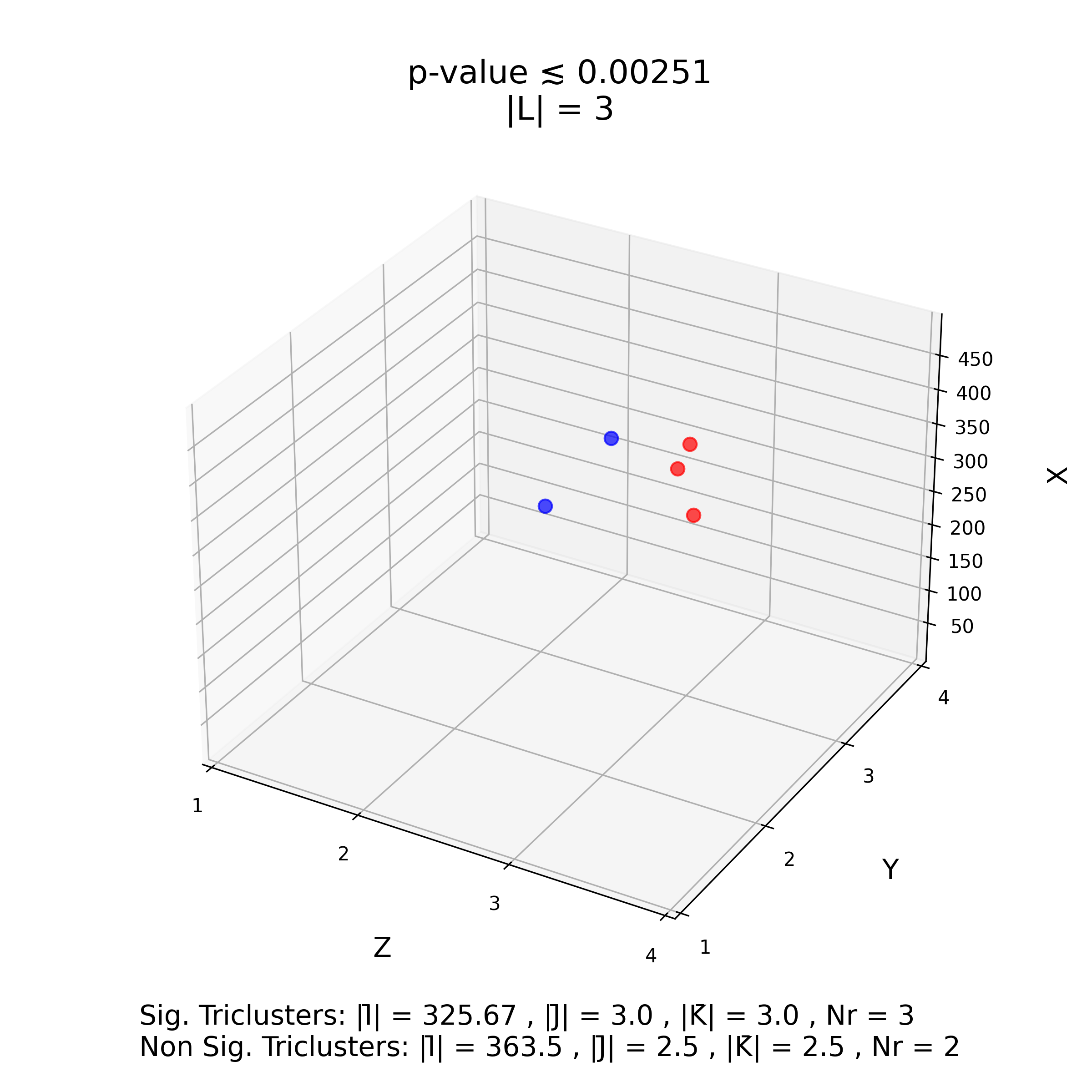}   
\includegraphics[height=2.5in]{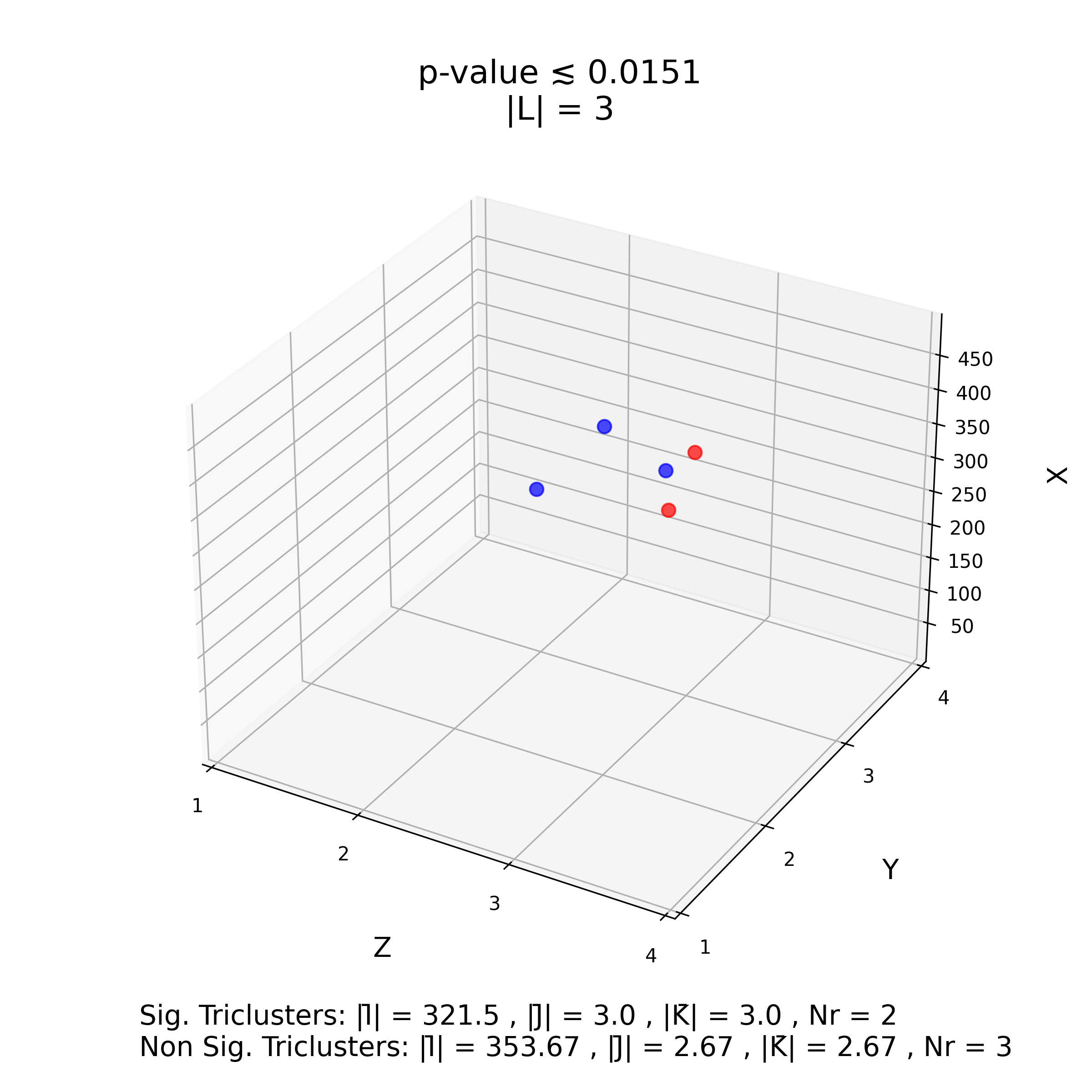}\\
\scriptsize{Without \textit{p}-value correction: \ \ \ \ \ \ \ \ \ \ \ \ \ \ \ \ \ \ \ \ \ \ \ \ With \textit{p}-value correction:} \\
\scriptsize{Variables Mutually dependent \ \ \ \ \ \ \ \ \ \ \ \ \ \ \ \ \ \ \ Variables Mutually dependent} \\
\scriptsize{Time-based Contexts \ \ \ \ \ \ \ \ \ \ \ \ \ \ \ \ \ \ \ \ \ \ \ \ \ \ \ \ \ \ \ Time-based Contexts} \\
\end{center}

\captionof{figure}{Results for the planted triclusters in synthetic data where variables either follow an Uniform or Gaussian distribution, and when considering varying dependency and time contiguity assumptions. 
Each dot on the scatter plot represents a tricluster. The position of the tricluster is defined by its number of observations ($|I|$), variables ($|J|$), and contexts ($|K|$). The dots are colored red if they are below the statistical significance threshold defined by the Benjamini-Hochberg procedure, colored orange if they are above the threshold but are below $0.05$, and colored blue otherwise. 
In the first column of scatter plots \textit{p}-value correction is not applied, in the second column \textit{p}-value correction is applied.}
\label{synthetic_datasets}
\end{adjustwidth}
\end{figure}

\begin{table}[!b]
\begin{adjustwidth}{-1cm}{0cm}

\begin{center}
    \captionof{table}{ \small Statistics of the triclusters presented in Figure \ref{synthetic_datasets} when variables follow either a uniform or Gaussian distribution. Column ``$(|I|, |J|, |K|)$" presents details pertaining to the volume of the triclusters. Column ``$p_{\varphi_{B_k}}$" presents intermediate statistics necessary to calculate $p_{\varphi_{T}}$.
    Column ``\textit{p}-value" and ``\textit{p}-value (with correction)" contain the \textit{p}-values obtained using the binomial test of significance, with and without the proposed correction.}
    \vspace{0.2cm}

\scriptsize
\begin{tabular}{c|ccc}\toprule
\multirow{2}{*}{\begin{tabular}[c]{@{}c@{}}Triclusters\\ ($|I|$, $|J|$, $|K|$)\end{tabular}} & \multicolumn{3}{c}{Synthetic data ($|X|$ = 1000, $|Y|$ = 50, $|Z||$ = 50) -- Uniform distribution}                                    \\ \cmidrule{2-4} 
                & \multicolumn{1}{c|}{$p_{\varphi_{T}} = {Z \choose k}\prod_{k \in K}^{} p_{\varphi_{B_k}}$} & \multicolumn{1}{c|}{\textit{p}-value}                                                                 & \textit{p}-value (with correction) \\ \midrule
\multirow{1}{*}{(55, 3, 2)}                                                                  & \multicolumn{1}{c|}{$p_{\varphi_{B_{k_1}}} \approx 5.2 \times 10^{-3}$}                          & \multicolumn{1}{c|}{\multirow{1}{*}{$2.9 \times 10^{-4}$}}                                   
& \multirow{1}{*}{$1.0$}    \\\midrule
\multirow{1}{*}{(484, 2, 2)}                                                                 & \multicolumn{1}{c|}{$p_{\varphi_{B_{k_1}}} \approx 0.04$}                                        & \multicolumn{1}{c|}{\multirow{1}{*}{$1.0$}}                                                  & \multirow{1}{*}{$1.0$}    \\\midrule
\multirow{1}{*}{(249, 3, 2)}                                                                 & \multicolumn{1}{c|}{$p_{\varphi_{B_{k_1}}} \approx 0.01$}                                        & \multicolumn{1}{c|}{\multirow{1}{*}{$8.9 \times 10^{-6}$}} & \multirow{1}{*}{$0.17$}   \\\midrule
\multirow{1}{*}{(327, 3, 2)}                                                                 & \multicolumn{1}{c|}{$p_{\varphi_{B_{k_1}}} \approx 0.02$}                                        & \multicolumn{1}{c|}{\multirow{1}{*}{$0.13$}}                                                 & \multirow{1}{*}{$1.0$}    \\\midrule
\multirow{1}{*}{(85, 2, 3)}                                                                  & \multicolumn{1}{c|}{$p_{\varphi_{B_{k_1}}} \approx 0.03$}                                        & \multicolumn{1}{c|}{\multirow{1}{*}{$1.0$}}                                                  & \multirow{1}{*}{$1.0$}    \\\bottomrule
\end{tabular}


\vskip 0.5cm

\scriptsize
\begin{tabular}{c|ccc}\toprule
\multirow{2}{*}{\begin{tabular}[c]{@{}c@{}}Triclusters\\ ($|I|$, $|J|$, $|K|$)\end{tabular}} & \multicolumn{3}{c}{Synthetic data ($|X|$ = 1000, $|Y|$ = 50, $|Z||$ = 50) -- Gaussian distribution}            \\ \cmidrule{2-4} 
                & \multicolumn{1}{c|}{$p_{\varphi_{T}} = (|Z| - |K| + 1) \times p_{\varphi_{B_{k_1}}} \times \prod_{k \in K \backslash k_1}^{} p_{\varphi_{B_k}|\varphi_{B_{k-1}}}$} & \multicolumn{1}{c|}{p-value}                                 & p-value (with correction)               \\ \midrule
\multirow{1}{*}{(368, 3, 3)}                                                                 & \multicolumn{1}{c|}{$p_{\varphi_{B_{k_1}}} = 0.11$, $p_{\varphi_{B_{k_2}}|\varphi_{B_{k_1}}} = 0.22$, $p_{\varphi_{B_{k_3}}|\varphi_{B_{k_2}}} = 0.22$}                                   & \multicolumn{1}{c|}{\multirow{1}{*}{$2.48 \times 10^{-14}$}} & \multirow{1}{*}{$4.87 \times 10^{-10}$} \\\midrule
\multirow{1}{*}{(498, 2, 3)}                                                                 & \multicolumn{1}{c|}{$p_{\varphi_{B_{k_1}}} = 0.24$, $p_{\varphi_{B_{k_2}}|\varphi_{B_{k_1}}} = 0.29$, $p_{\varphi_{B_{k_3}}|\varphi_{B_{k_2}}} = 0.29$}                                   & \multicolumn{1}{c|}{\multirow{1}{*}{$1.0$}}                  & \multirow{1}{*}{$1.0$}                  \\\midrule
\multirow{1}{*}{(334, 3, 3)}                                                                 & \multicolumn{1}{c|}{$p_{\varphi_{B_{k_1}}} = 0.12$, $p_{\varphi_{B_{k_2}}|\varphi_{B_{k_1}}} = 0.22$, $p_{\varphi_{B_{k_3}}|\varphi_{B_{k_2}}} = 0.22$}                                   & \multicolumn{1}{c|}{\multirow{1}{*}{$0.003$}}                & \multirow{1}{*}{$1.0$}                  \\\midrule
\multirow{1}{*}{(275, 3, 3)}                                                                 & \multicolumn{1}{c|}{$p_{\varphi_{B_{k_1}}} = 0.11$, $p_{\varphi_{B_{k_2}}|\varphi_{B_{k_1}}} = 0.20$, $p_{\varphi_{B_{k_3}}|\varphi_{B_{k_2}}} = 0.20$}                                   & \multicolumn{1}{c|}{\multirow{1}{*}{$7.7 \times 10^{-7}$}}   & \multirow{1}{*}{$0.02$}                 \\\midrule
\multirow{1}{*}{(229, 3, 2)}                                                                 & \multicolumn{1}{c|}{$p_{\varphi_{B_{k_1}}} = 0.12$, $p_{\varphi_{B_{k_2}}|\varphi_{B_{k_1}}} = 0.17$}                                   & \multicolumn{1}{c|}{\multirow{1}{*}{$1.0$}}                  & \multirow{1}{*}{$1.0$}                  \\\bottomrule
\end{tabular}
\label{synthetic_gaussian_table_time}
\end{center}

\end{adjustwidth}
\end{table}

Consider the results for the synthetic data presented in Figure \ref{synthetic_datasets}. Assuming variables and contexts to be mutually dependent, we selected two cases: 1) planted triclusters are not contiguous in the context dimension, and variables follow a uniform distribution, and 2) triclusters are contiguous in the context dimension, and variables follow a Gaussian distribution. The pattern characteristics, intermediate statistics, and \textit{p}-values are presented in Table 
\ref{synthetic_gaussian_table_time}. By observing the aforesaid figure and table, the randomly planted triclusters with statistical significance are not only conditioned  by the number of observations, variables, and contexts they occur, but as well by the probability of the tricluster pattern, $\varphi_{T}$. 
In this case, with mutually dependent contexts and no context contiguity, the joint probability of all slice patterns,  $\varphi_{B}$, contained in $\varphi_{T}$ is balanced by ${Z \choose k}$. If patterns $\varphi_{B}$ are not sufficiently improbable in their corresponding variables, meaning they don't occur in a high number of contexts or are probable to occur, the pattern will not yield statistical significance. We observe this in triclusters $(|I| = 55, |J| = 3, |K| = 2)$ and $(|I| = 85, |J| = 2, |K| = 3)$. 
Despite both tricluster patterns occurring in a low number of observations, the pattern is not rare enough to be considered statistically significant. 
Considering the Gaussian setting, where contexts are mutually dependent and triclusters contiguous across contexts, the joint probability between $\varphi_{B_{k_1}}$ and $\varphi_{T}$ transitions, and a more relaxed statistic $(|Z| - |K| + 1)$, allow more probable patterns to have statistical significance. 
Tricluster patterns with a higher number of variables and contexts yield statistical significance. 
Finally, the changes observed under the \textit{p}-value correction suggest the importance of this procedure to 
effectively mitigate possible false-positives.

\section{Conclusion}

This work proposed a novel methodology to rigorously assess the statistical significance of patterns extracted from tensor data. The methodology provides a robust set of statistical principles that accommodate different aspects of tensor data such as variable domains and dependencies, temporal dependencies and misalignments, 
and relevant \textit{p}-value corrections. These principles can be easily extended to $N$-way tensor data, bridging the current gap on how to assess the statistical significance of patterns in tensor data with more than two dimensions. Results gathered from real-world case studies provide notable validity and in-depth reasoning on how to apply the methodology in practice. 

The implementation of the proposed statistical tests can be found at \href{https://github.com/JupitersMight/TriSig}{https:// github.com/JupitersMight/TriSig}. These statistical tests can be integrated into the discovery process of patterns, aiding in pattern-centric descriptive and predictive tasks. 

\subsection*{Funding}
\small\noindent This work was supported by the FCT PhD grant to LA (2021.07759.BD), Associate Laboratory for Green Chemistry (LAQV) financed by national funds from FCT/MCTES (UIDB/50006/2020 and UIDP/50006/2020), INESC-ID plurianual (UIDB/50021/2020), and the contract CEECIND/01399/2017 to RSC. The authors also wish to acknowledge the European Union’s Horizon BioLaMer project under grant agreement number [101099487].




\footnotesize
\bibliographystyle{elsarticle-harv} 
\bibliography{references}
\end{document}